\documentclass[10pt,twocolumn,letterpaper]{article}

\usepackage[pagenumbers]{cvpr} 

\usepackage{tcolorbox}
\usepackage{tabularx}
\usepackage{pifont}

\usepackage{tcolorbox}
\usepackage{tabularx}
\usepackage{colortbl}

\usepackage{pgfplots}
\usepackage{tikz}
\pgfplotsset{compat=1.18}
\usepgfplotslibrary{polar}

\usepackage{listings}
\usepackage{enumitem}
\usepackage[accsupp]{axessibility}

\usepackage{graphicx}
\usepackage{subcaption}
\usepackage{tcolorbox}
\usepackage{xcolor}

\definecolor{cvprblue}{rgb}{0.21,0.49,0.74}
\usepackage[pagebackref,breaklinks,colorlinks,allcolors=cvprblue]{hyperref}

\def\paperID{*****} 
\def\confName{CVPR}
\def\confYear{2025}

\title{ReasonDrive: Efficient Visual Question Answering for Autonomous Vehicles with Reasoning-Enhanced Small Vision-Language Models}

\author{Amirhosein Chahe ~~~~~~ Lifeng Zhou\thanks{Corresponding author.}\\
Drexel University\\
Philadelphia PA 19104, USA\\
{\tt\small \{ac4462, lz457\}@drexel.edu}
}

\begin{document}
\maketitle
\begin{abstract}
Vision-language models (VLMs) show promise for autonomous driving but often lack transparent reasoning capabilities that are critical for safety. We investigate whether explicitly modeling reasoning during fine-tuning enhances VLM performance on driving decision tasks. Using GPT-4o, we generate structured reasoning chains for driving scenarios from the DriveLM benchmark with category-specific prompting strategies. We compare reasoning-based fine-tuning, answer-only fine-tuning, and baseline instruction-tuned models across multiple small VLM families (Llama 3.2, Llava 1.5, and Qwen 2.5VL). Our results demonstrate that reasoning-based fine-tuning consistently outperforms alternatives, with Llama3.2-11B-reason achieving the highest performance. Models fine-tuned with reasoning show substantial improvements in accuracy and text generation quality, suggesting explicit reasoning enhances internal representations for driving decisions. These findings highlight the importance of transparent decision processes in safety-critical domains and offer a promising direction for developing more interpretable autonomous driving systems. The code and dataset is available at \href{https://github.com/Zhourobotics/ReasonDrive}{https://github.com/Zhourobotics/ReasonDrive}.
\end{abstract}


\begin{figure*}[t]
    \centering
    \begin{minipage}{\textwidth}
        \centering
        \begin{subfigure}[b]{0.32\textwidth}
            \includegraphics[width=\textwidth]{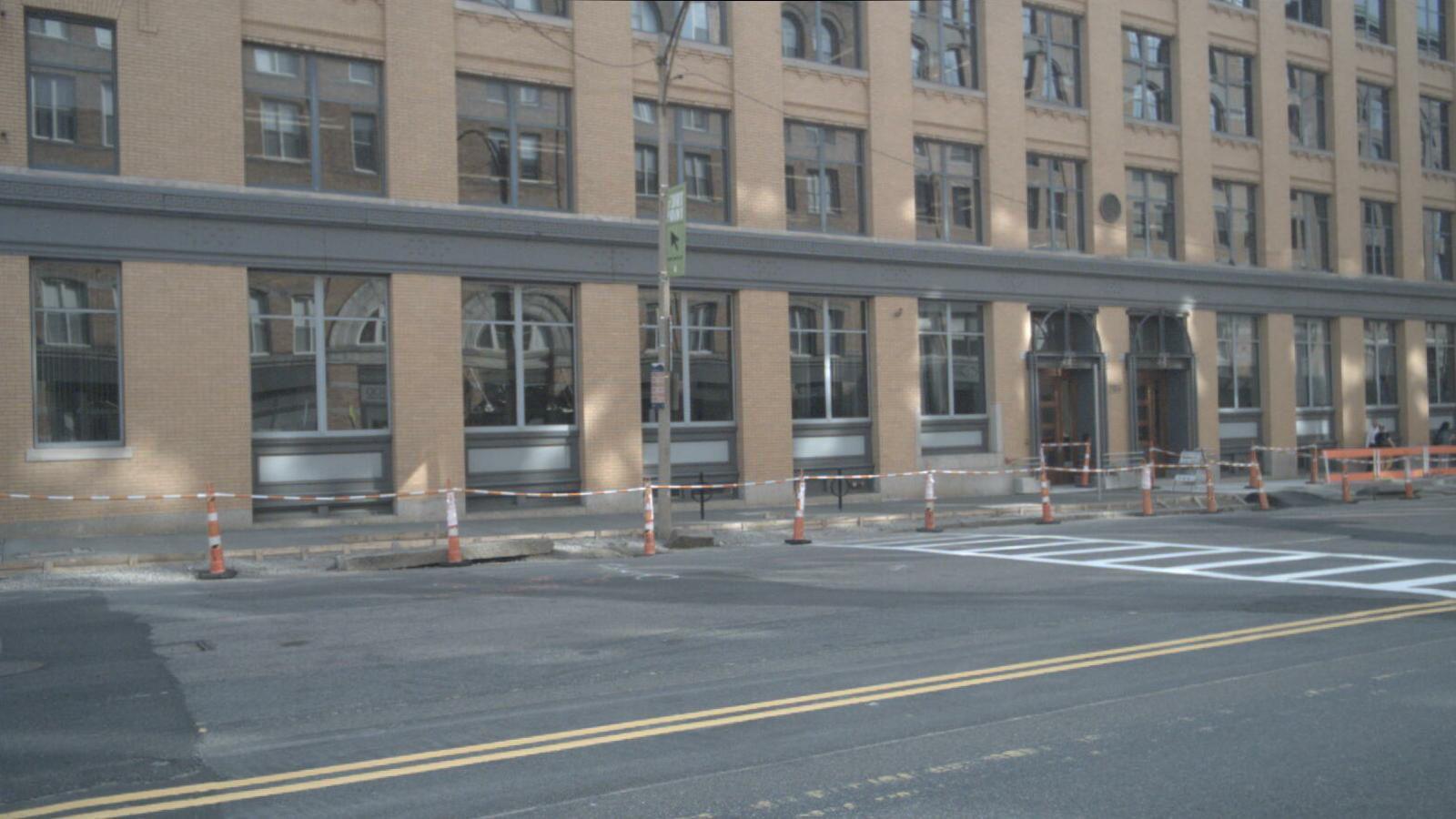}
            \caption{Front Left}
            \label{fig:cam_front_left}
        \end{subfigure}
        \hfill
        \begin{subfigure}[b]{0.32\textwidth}
            \includegraphics[width=\textwidth]{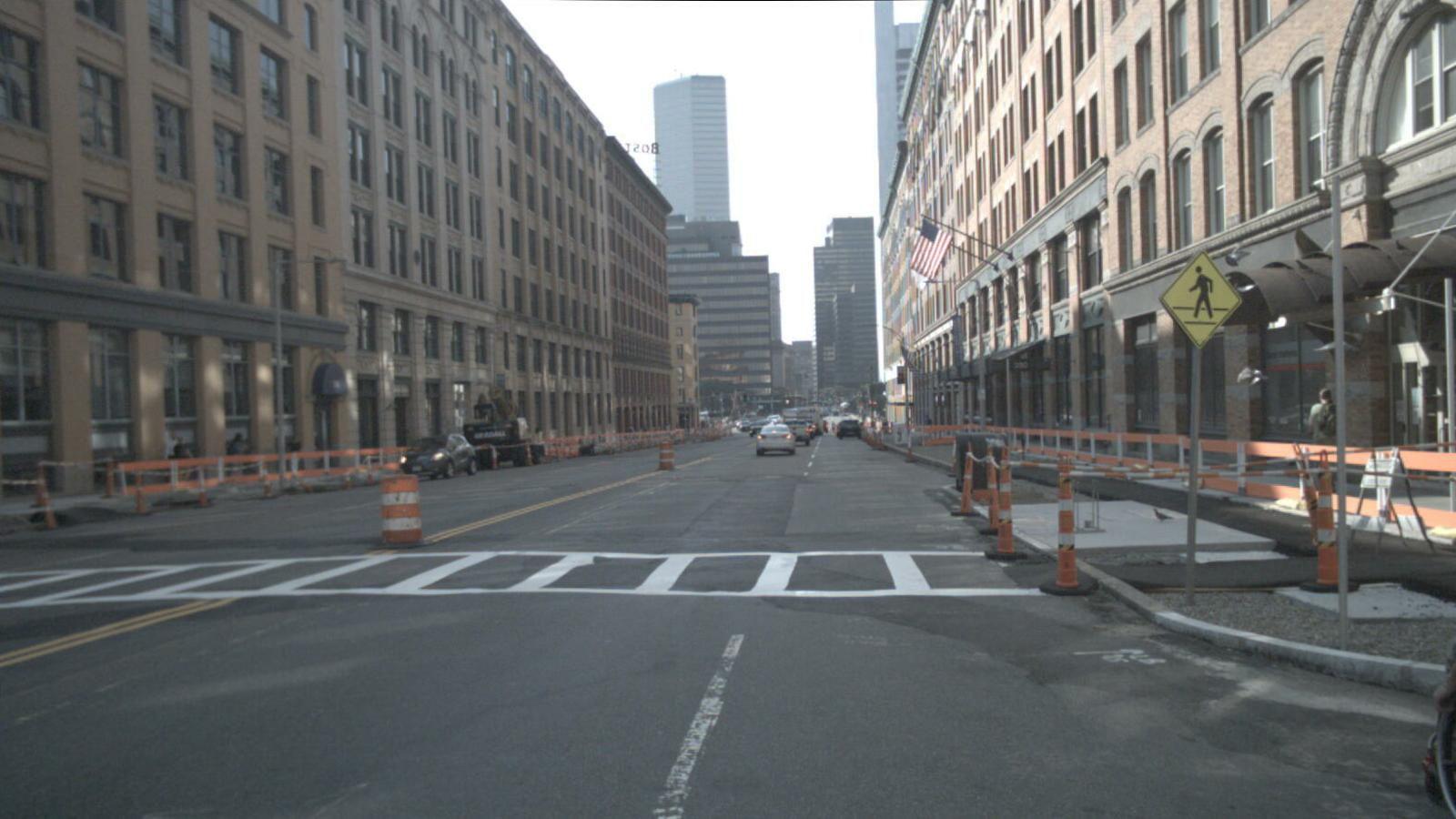}
            \caption{Front}
            \label{fig:cam_front}
        \end{subfigure}
        \hfill
        \begin{subfigure}[b]{0.32\textwidth}
            \includegraphics[width=\textwidth]{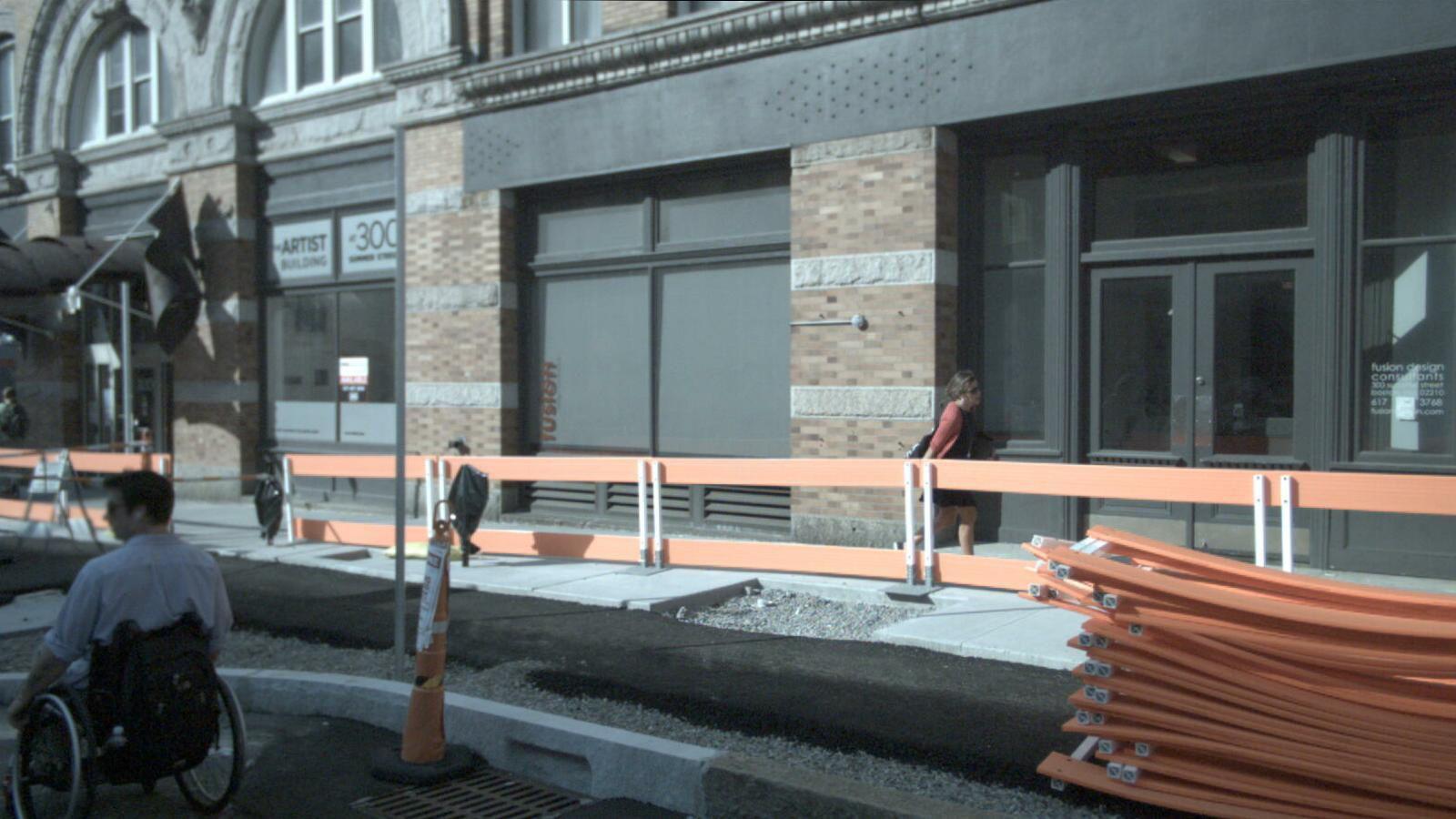}
            \caption{Front Right}
            \label{fig:cam_front_right}
        \end{subfigure}
        
        \vspace{0.05cm}
        
        \begin{subfigure}[b]{0.32\textwidth}
            \includegraphics[width=\textwidth]{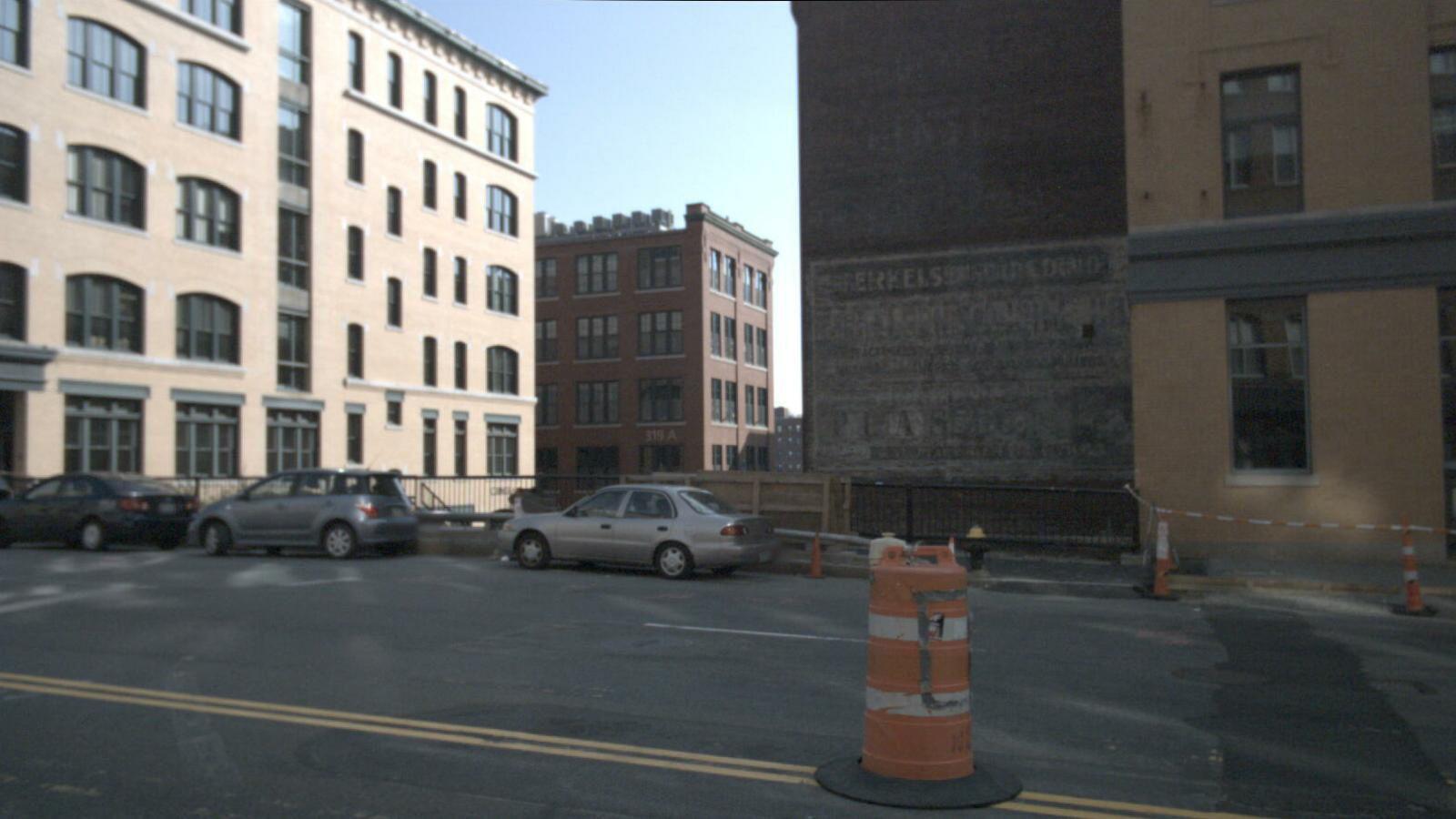}
            \caption{Back Left}
            \label{fig:cam_back_left}
        \end{subfigure}
        \hfill
        \begin{subfigure}[b]{0.32\textwidth}
            \includegraphics[width=\textwidth]{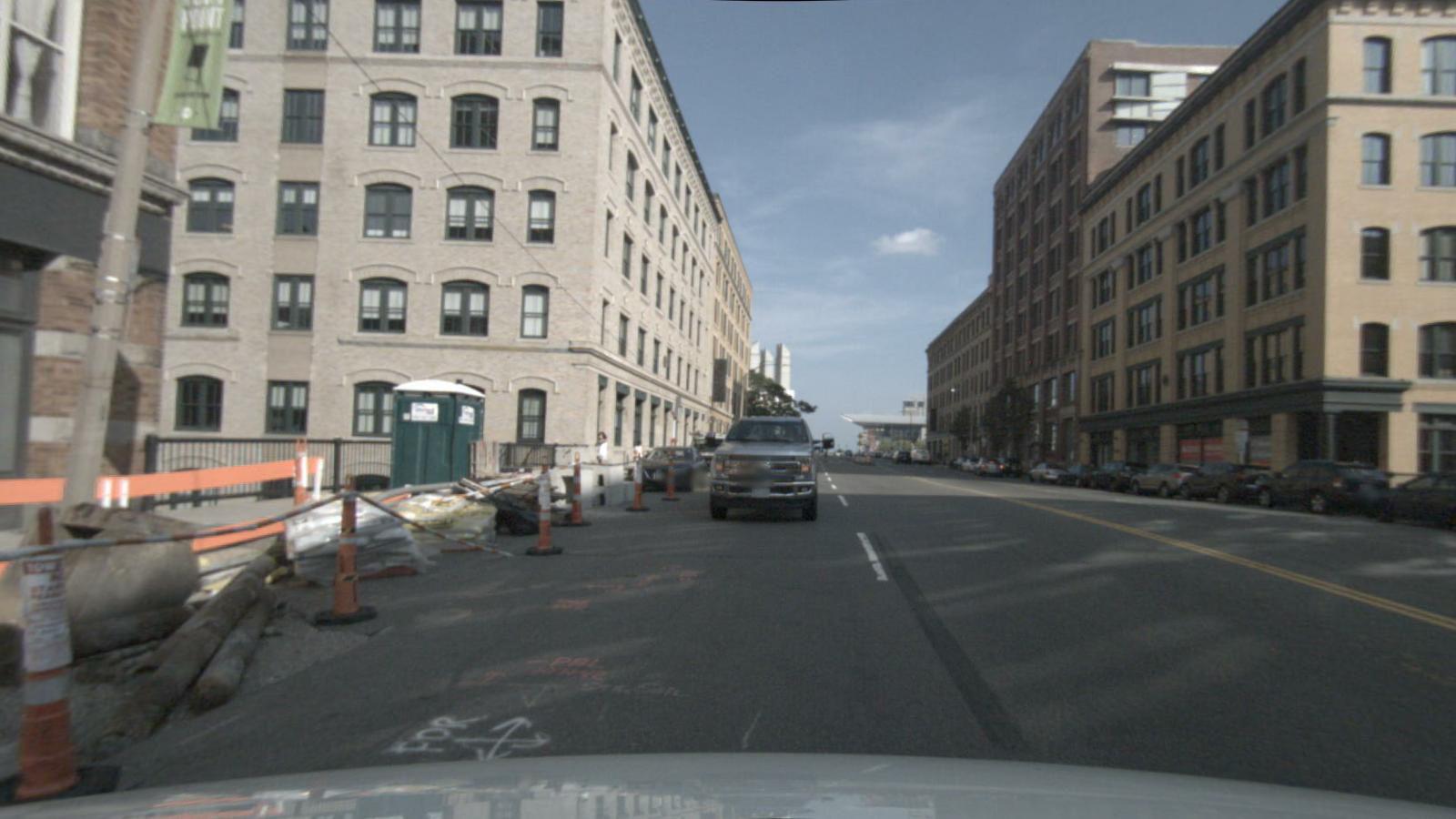}
            \caption{Back}
            \label{fig:cam_back}
        \end{subfigure}
        \hfill
        \begin{subfigure}[b]{0.32\textwidth}
            \includegraphics[width=\textwidth]{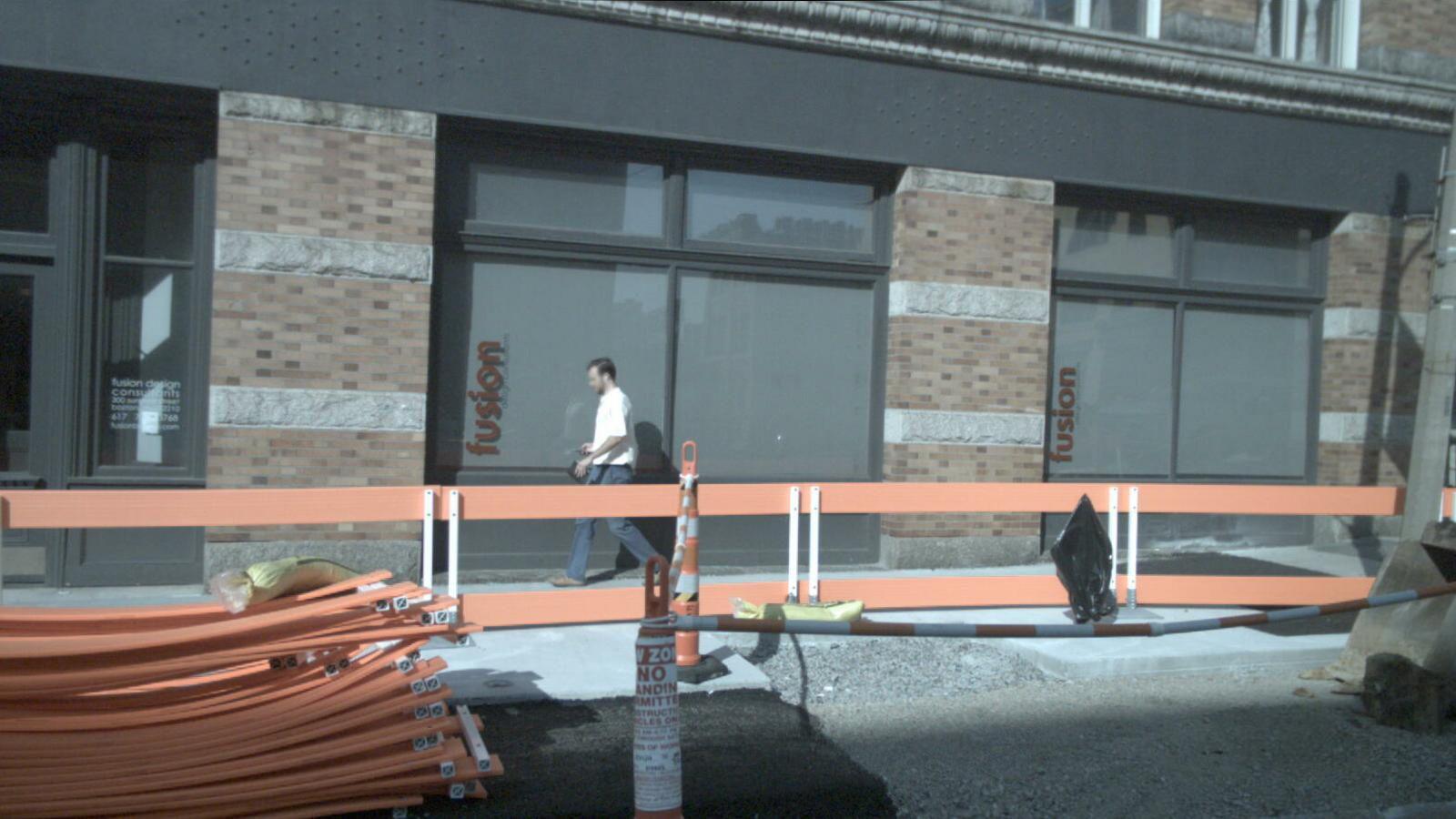}
            \caption{Back Right}
            \label{fig:cam_back_right}
        \end{subfigure}
    \end{minipage}
    
    \vspace{0.1cm}
    
    \begin{minipage}{\textwidth}
        \scriptsize 
        \renewcommand{\baselinestretch}{0.8} 
        \begin{tcolorbox}[colback=blue!5, colframe=blue!30, title=\textbf{Perception}, fonttitle=\bfseries]
            \textbf{Question:} What are the important objects in the current scene? Those objects will be considered for the future reasoning and driving decision. \\
            \textbf{Thinking:} Observe the scene and note a busy urban environment and construction on the sides. Identify key objects, focusing on the positions relative to the ego vehicle: a black car and a white sedan ahead, a silver SUV behind, and a person in a wheelchair to the front right. Note visual cues such as the pedestrian sign indicating yielding to pedestrians in front of the ego vehicle. Input the identified objects and their positions as crucial for future driving decisions. \\
            \textbf{Answer:} There is a black car to the front of the ego vehicle, a white sedan to the front of the ego vehicle, a silver SUV to the back of the ego vehicle, a person in a wheelchair to the front right of the ego vehicle, and a yield to pedestrians to the front of the ego vehicle. The IDs of these objects are $<c1>$, $<c2>$, $<c3>$, $<c4>$, and $<c5>$.
        \end{tcolorbox}
    \end{minipage}
    
    
    \begin{minipage}{\textwidth}
        \scriptsize
        \renewcommand{\baselinestretch}{0.8}
        \begin{tcolorbox}[colback=green!5, colframe=green!30, title=\textbf{Prediction}, fonttitle=\bfseries]
            \textbf{Question:} Is $<c1>$ a traffic sign or a road barrier? \\
            \textbf{Thinking:} Analyze the location and characteristics of the object in question based on its coordinates $<c1>$ relative to the camera. Determine if the object possesses common features of a traffic sign, such as specific shapes, colors, or placement above ground level. Check if the object resembles typical road barriers, which are usually positioned at ground level, often elongated, and can be movable or fixed. Given the data input, evaluate whether it matches the expected attributes of either a traffic sign or a road barrier. The identification process shows no match for typical features of traffic signs or road barriers; thus, conclude the object is neither. \\
            \textbf{Answer:} No.
        \end{tcolorbox}
    \end{minipage}
    
    
    \begin{minipage}{\textwidth}
        \scriptsize
        \renewcommand{\baselinestretch}{0.8}
        \begin{tcolorbox}[colback=orange!5, colframe=orange!30, title=\textbf{Planning}, fonttitle=\bfseries]
            \textbf{Question:} In this scenario, what are safe actions to take for the ego vehicle? \\
            \textbf{Thinking:} The scenario involves navigating a street partially obstructed by construction. Safety assessment dictates prioritizing the avoidance of sudden braking to maintain stability and control, especially around construction areas. Slightly offsetting to the left reduces the risk of collision with obstacles on the right without veering into oncoming traffic, optimizing lane position. Decelerating gradually ensures the vehicle adapts to the changing environment smoothly, minimizing the risk of losing control or causing discomfort to passengers. This approach balances maintaining safety and ensuring efficient movement through the construction zone. \\
            \textbf{Answer:} Slightly offset to the left, decelerate gradually without braking. Slightly offset to the left, decelerate gradually without braking.
        \end{tcolorbox}
    \end{minipage}
    
    
    \begin{minipage}{\textwidth}
        \scriptsize
        \renewcommand{\baselinestretch}{0.8}
        \begin{tcolorbox}[colback=purple!5, colframe=purple!30, title=\textbf{Behavior}, fonttitle=\bfseries]
            \textbf{Question:} Predict the behavior of the ego vehicle. \\
            \textbf{Thinking:} 1) Identify the road structure: The images show a straight road with lane markings directing vehicles to continue straight.
            2) Consider environmental factors: Construction cones and pedestrian signs indicate a need for caution, likely leading to slower speeds. 3) Analyze motion pattern and trajectory: Observing the surrounding vehicles and lane markings implies the ego vehicle is maintaining its straight trajectory. 4) Determine intent based on context: Due to construction and pedestrian considerations, the ego vehicle is likely driving slowly to ensure safety. \\
            \textbf{Answer:} The ego vehicle is going straight. The ego vehicle is driving slowly.
        \end{tcolorbox}
    \end{minipage}

    \caption{Example frame from our reasoning-enhanced dataset derived from DriveLM. Each frame consists of six camera views (top) and four categories of driving tasks (bottom) with the original question, generated reasoning, and final answer.}
    \label{fig:dataset_example}
\end{figure*}

\section{Introduction}
\label{sec:intro}

Visual Question Answering (VQA) represents a critical advancement in autonomous driving technology, enabling vehicles to understand and respond to natural language questions about their visual surroundings. This capability bridges the gap between computer vision and natural language processing, allowing autonomous vehicles to interpret road scenes and communicate their understanding in human-readable formats \cite{Rekanar2023, chahe2025query3d}. As a safety-critical application, autonomous driving necessitates explainable decision-making, which VQA can facilitate through question-answering-based causal reasoning \cite{Atakishiyev2023}.

Applying VQA to autonomous driving presents unique challenges, including processing multi-modal data, analyzing multi-frame sequences from continuous real-time acquisition, and interpreting complex outdoor scenes containing both moving and static elements \cite{Qian2023}. To address these complexities, researchers have developed specialized datasets and benchmarks, such as NuScenes-QA, which encompasses 34,000 visual scenes and 460,000 question-answer pairs specifically designed for autonomous driving scenarios \cite{Qian2023}.

The integration of VQA in autonomous vehicles serves multiple purposes beyond basic perception. It enables natural interaction between the vehicle and its occupants, fosters trust in autonomous technology, and makes the vehicle's decision-making process more transparent \cite{Rekanar2024}. This transparency is crucial for building user confidence in autonomous systems, as it allows vehicles to communicate their actions, intentions, and reasoning in human-comprehensible language.
Recent large vision-language models (VLMs)~\cite{GPT4o, gemini, claude3} offer promising alternatives to traditional domain-specific approaches, potentially performing multiple perception tasks within a unified framework \cite{Dewangan2023}. However, deploying these large models in vehicles presents significant challenges due to computational constraints and the need for interpretable reasoning. While benchmarks like DriveLM \cite{Sima2023} have advanced VLM applications in driving scenarios, a critical gap remains in making these capabilities accessible within real-world computational constraints without sacrificing reasoning transparency. Current approaches often rely on massive foundation models impractical for vehicle deployment or compromise on transparency for efficiency~\cite{Cui_2024_WACV}.

We address this gap by investigating whether explicitly modeling reasoning during fine-tuning enhances smaller, deployable VLMs on driving decision tasks. Our contributions include:

\begin{enumerate}
    \item A reasoning-enhanced dataset derived from DriveLM using GPT-4o to generate structured reasoning chains for driving scenarios.
    
    \item Comprehensive evaluation of reasoning-based fine-tuning across multiple small VLM families (Llama 3.2, Llava 1.5, and Qwen 2.5VL), demonstrating that reasoning-based approaches consistently outperform alternatives.
    
    \item Evidence that explicit reasoning enhances internal representations for driving decisions, with Llama3.2-11B-reason achieving the highest overall performance while maintaining interpretability.
\end{enumerate}

Our work shows that reasoning-enhanced fine-tuning creates efficient, interpretable models that address both computational and safety requirements for autonomous vehicles. Figure~\ref{fig:dataset_example} presents the structure of reasoning and corresponding answers for an example frame from the DriveLM dataset.

\section{Related Work}

\subsection{VQA and VLMs in Autonomous Driving}

VQA technologies for autonomous driving bridge vision and language to interpret complex road scenes \cite{Rekanar2023, Atakishiyev2023}. Key advancements include specialized datasets like NuScenes-QA with 34,000 visual scenes and 460,000 question-answer pairs \cite{Qian2023} and DriveLM's Graph VQA approach, which structures driving reasoning as interconnected question-answer pairs mimicking human multi-step reasoning \cite{Sima2023, Sheng2025}. Other notable contributions include LingoQA, featuring over 419,000 QA pairs focused on reasoning and action justifications \cite{Marcu2023}, and frameworks integrating VQA models with specialized visual perception modules \cite{He2024}. Recent frameworks structure autonomous driving tasks—perception, prediction, planning, and behavior—as series of VQA interactions \cite{Zheng2024, Liu2023, Wang2024, Zhou2025}. Large language and vision-language models have shown promise as alternatives to traditional domain-specific approaches \cite{Dewangan2023}, with architectures typically involving separate encoding of vision and text features, later fused and processed by LLMs \cite{Gopalkrishnan2024, Fourati2024, Guo2024}.

\subsection{DriveLM Dataset and Capabilities}

DriveLM introduces Graph Visual Question Answering for autonomous driving, recognizing that human drivers reason in multiple steps rather than single-round interactions \cite{Sima2023}. The dataset comprises DriveLM-nuScenes and DriveLM-CARLA with question-answer pairs arranged in graph structures linking visual perception with driving behaviors through logical reasoning. DriveLM covers the entire autonomous driving pipeline through three interconnected reasoning stages \cite{Sheng2025, Sima2023} and introduces DriveLM-Agent, which employs a trajectory tokenizer with graph prompting that models logical dependencies \cite{Sima2023, Li2023, Liu2023}. This approach effectively repurposes VLMs for autonomous driving, demonstrating competitive performance and enhanced generalization to unseen objects or sensor configurations \cite{Chen2024, Sima2023, Zhou2025}.

\subsection{Efficiency Techniques and Knowledge Distillation}

Deploying VLMs in vehicles faces challenges due to substantial computational requirements \cite{Gopalkrishnan2024}, as current research predominantly uses billion-parameter models requiring expensive hardware and longer inference times \cite{Gopalkrishnan2024, Zhang2024}.  To address these constraints, researchers have developed efficiency techniques including: lightweight architectures like EM-VLM4AD \cite{Zhang2024}, video token sparsification for reduced memory usage \cite{Ma2024}, parameter-efficient fine-tuning with LoRA \cite{Gopalkrishnan2024}, and dual architecture approaches \cite{Fourati2024}. Knowledge distillation frameworks have shown promising results, with gains of 21.40\% to 32.28\% across metrics on the DriveLM dataset \cite{Lin2025}.

Despite these advances, existing approaches typically either (1) use computationally impractical large models, (2) sacrifice reasoning transparency for efficiency, or (3) develop specialized architectures potentially lacking foundation model robustness. A significant gap remains in developing smaller, deployable VLMs that maintain interpretable reasoning while achieving performance comparable to larger models—a gap our work directly addresses by investigating reasoning-based fine-tuning for smaller VLMs as a practical pathway for real-world deployment in autonomous vehicles.

\begin{table}[ht]
\centering
\caption{Category-specific prompting strategy for generating reasoning chains}
\begin{tabular}{|p{1.5cm}|p{6.5cm}|}
\hline
\textbf{Category} & \textbf{Focus Areas and Instructions} \\
\hline
Perception & 
\begin{minipage}[t]{\linewidth}
\vspace{-0.4em}
\begin{itemize}[nosep,leftmargin=*,topsep=0pt]
\item Quickly summarize the observed scene
\item Identify key objects and their positions
\item Note immediate visual cues and statuses
\item Format response within \texttt{<think>} tags using 1 concise sentence
\end{itemize}
\end{minipage} \\
\hline
Prediction & 
\begin{minipage}[t]{\linewidth}
\vspace{-0.4em}
\begin{itemize}[nosep,leftmargin=*,topsep=0pt]
\item Concisely forecast future states based on current data
\item Consider object motion, momentum, and interactions
\item Apply basic traffic rules and driver behavior
\item Format response within \texttt{<think>} tags using 1-2 sentences
\end{itemize}
\end{minipage} \\
\hline
Planning & 
\begin{minipage}[t]{\linewidth}
\vspace{-0.4em}
\begin{itemize}[nosep,leftmargin=*,topsep=0pt]
\item Assess safety and prioritize actions
\item Evaluate decision options and trade-offs
\item Consider alternative actions and consequences
\item Format response within \texttt{<think>} tags using 2-3 sentences
\end{itemize}
\end{minipage} \\
\hline
Behavior & 
\begin{minipage}[t]{\linewidth}
\vspace{-0.4em}
\begin{itemize}[nosep,leftmargin=*,topsep=0pt]
\item Analyze motion patterns, speed, and trajectories
\item Consider environmental factors and multi-view observations
\item Determine the underlying intent based on dynamic context
\item Format response within \texttt{<think>} tags using 1-2 concise sentences
\end{itemize}
\end{minipage} \\
\hline
\end{tabular}
\label{tab:prompting_strategy}
\end{table}

\section{Methodology}

\subsection{Reasoning-Enhanced VQA}
To investigate the impact of explicit reasoning on driving decision tasks, we developed a reasoning-enhanced approach for visual question answering in autonomous driving scenarios. Our approach builds upon the DriveLM dataset~\cite{Sima2023}, enhancing it with structured reasoning chains that explain the decision-making process.

We prompted GPT-4o~\cite{GPT4o} with multicamera inputs, the original questions, and the ground truth answers, requesting it to generate structured reasoning that explains the steps to the provided answer. Our prompting strategy was tailored to each driving task category as shown in Table~\ref{tab:prompting_strategy}. This category-specific prompting ensured that the generated reasoning aligned with the particular requirements of each driving task. All prompts began with a common system prompt: ``You are assisting in developing a reasoning system for an autonomous driving AI. Below is a question and answer pair from the `[category]' category. Your task is to generate a structured, step-by-step reasoning process that logically leads to the provided answer.''

After obtaining the reasoning outputs from GPT-4o, we structured each training example with the following components: (1) a system prompt for driving assistance, (2) the original question, (3) the generated reasoning, (4) the ground truth answer wrapped in \texttt{<answer>} tags, (5) the six camera view images, and (6) metadata including the original question-answer pair, scene ID, and frame ID. The resulting dataset contains the original questions and answers, enhanced with explicit reasoning that connects visual observations to driving decisions. Figure~\ref{fig:dataset_example} depicts an example frame from the dataset.

\subsection{Fine-Tuning with Reasoning}
We propose a reasoning-based fine-tuning approach for vision-language models (VLMs) on driving tasks. This approach contrasts with traditional fine-tuning methods that directly optimize for the answer without explicitly modeling the reasoning process. We aim to improve model decision-making transparency and performance by incorporating structured reasoning.
During training, models learn to first generate explicit reasoning and then produce the final answer.
During inference, models are prompted to provide their thinking within \texttt{<think>} tags before answering within \texttt{<answer>} tags.

\begin{figure*}[ht]
\centering

    \begin{minipage}{\textwidth}
        \centering
        \begin{subfigure}[b]{\textwidth}
    \centering
    \begin{minipage}{\textwidth}
        \centering
        \begin{subfigure}[b]{0.32\textwidth}
            \includegraphics[width=\textwidth,height=2.6cm]{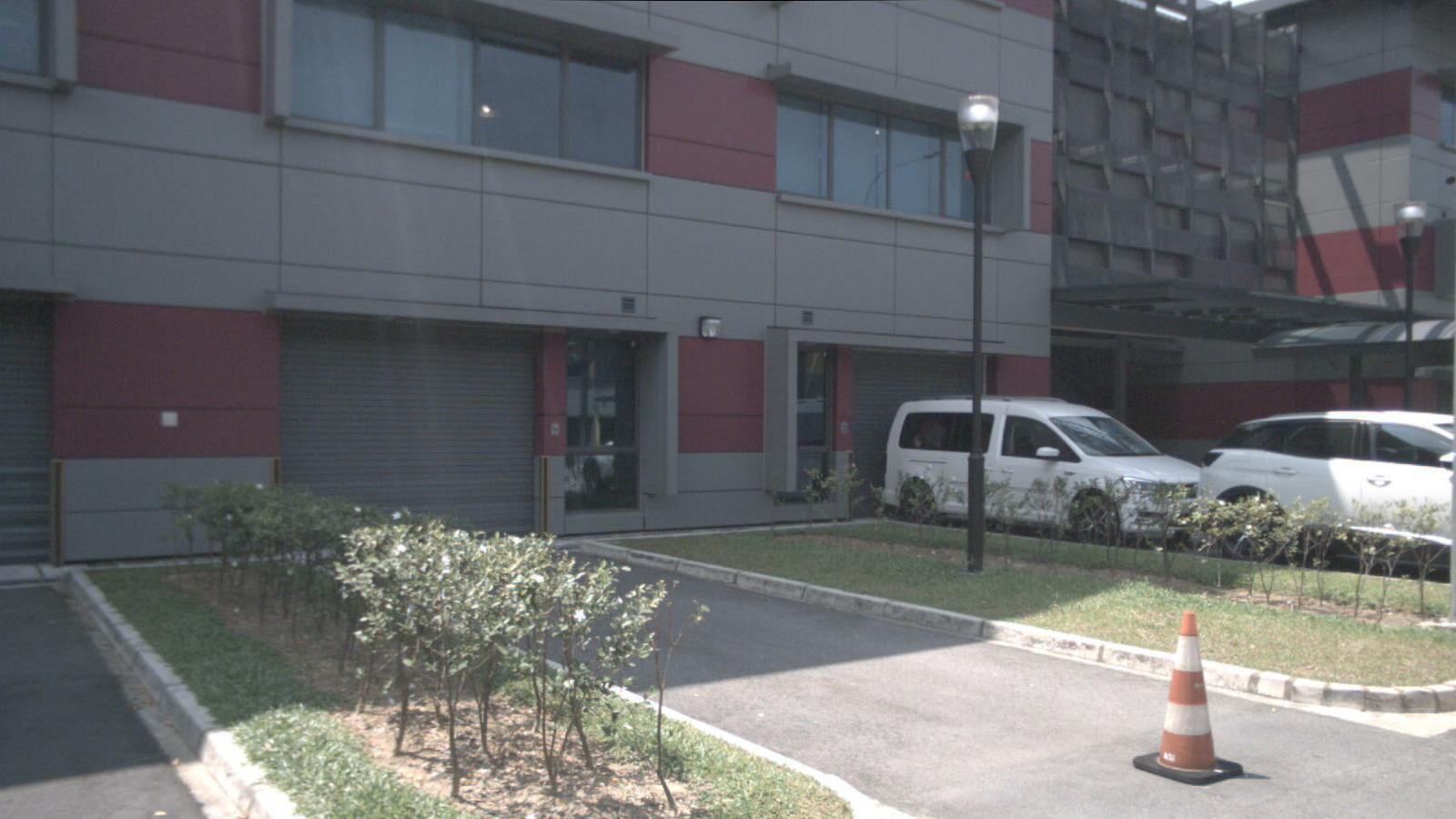}
        \end{subfigure}
        \hfill
        \begin{subfigure}[b]{0.32\textwidth}
            \includegraphics[width=\textwidth,height=2.6cm]{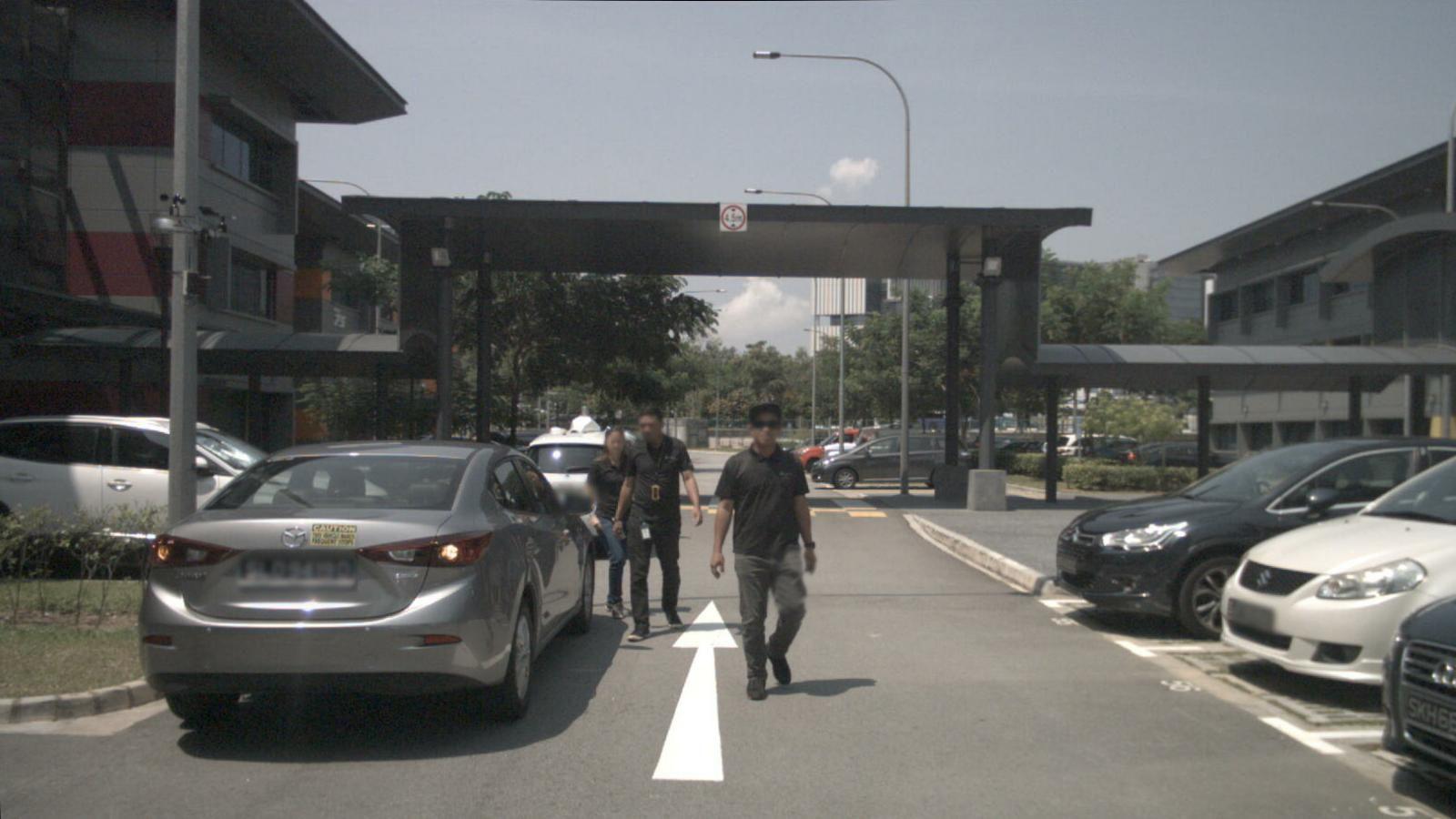}
        \end{subfigure}
        \hfill
        \begin{subfigure}[b]{0.32\textwidth}
            \includegraphics[width=\textwidth,height=2.6cm]{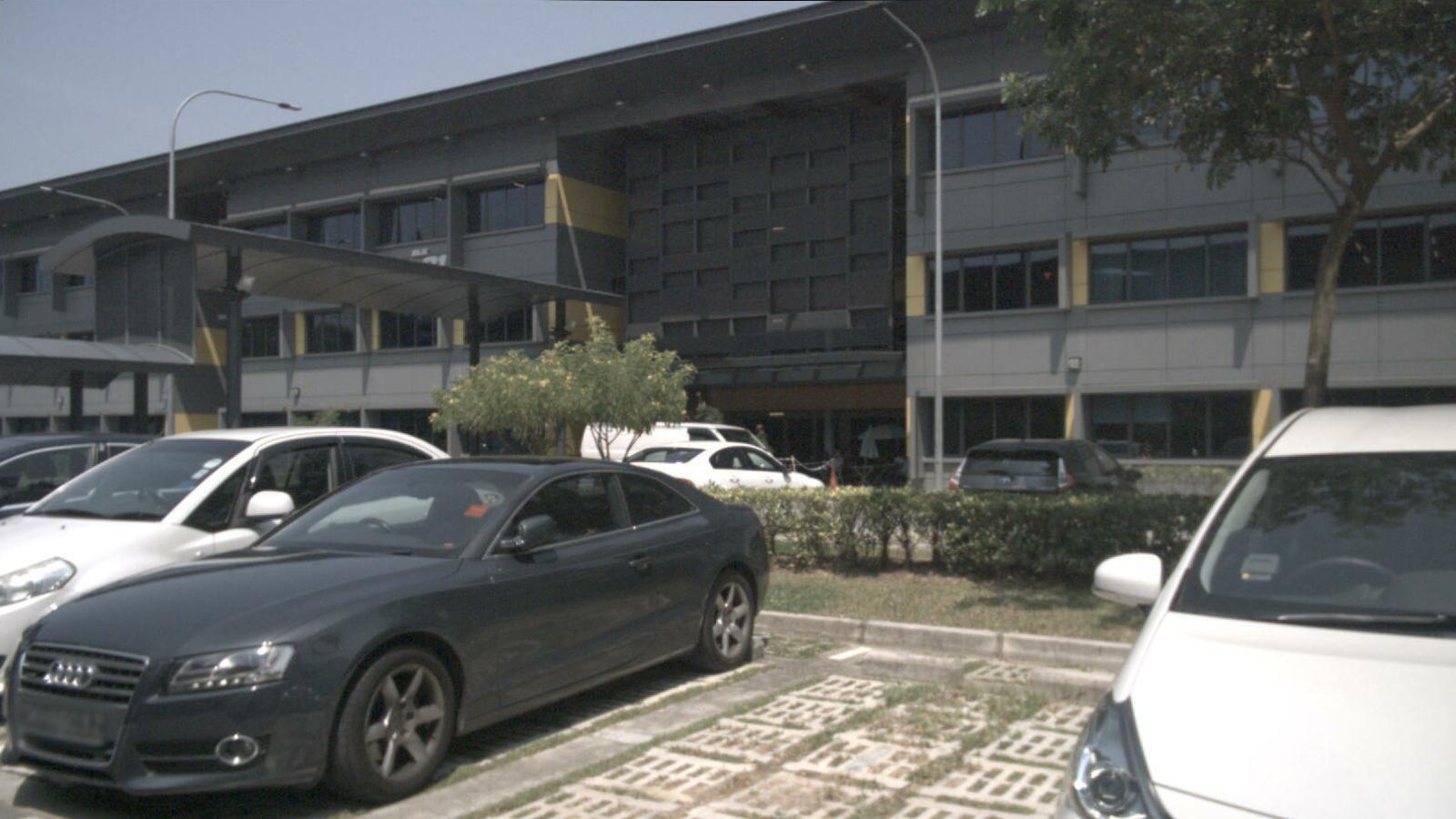}
        \end{subfigure}
        
        
        \begin{subfigure}[b]{0.32\textwidth}
            \includegraphics[width=\textwidth,height=2.6cm]{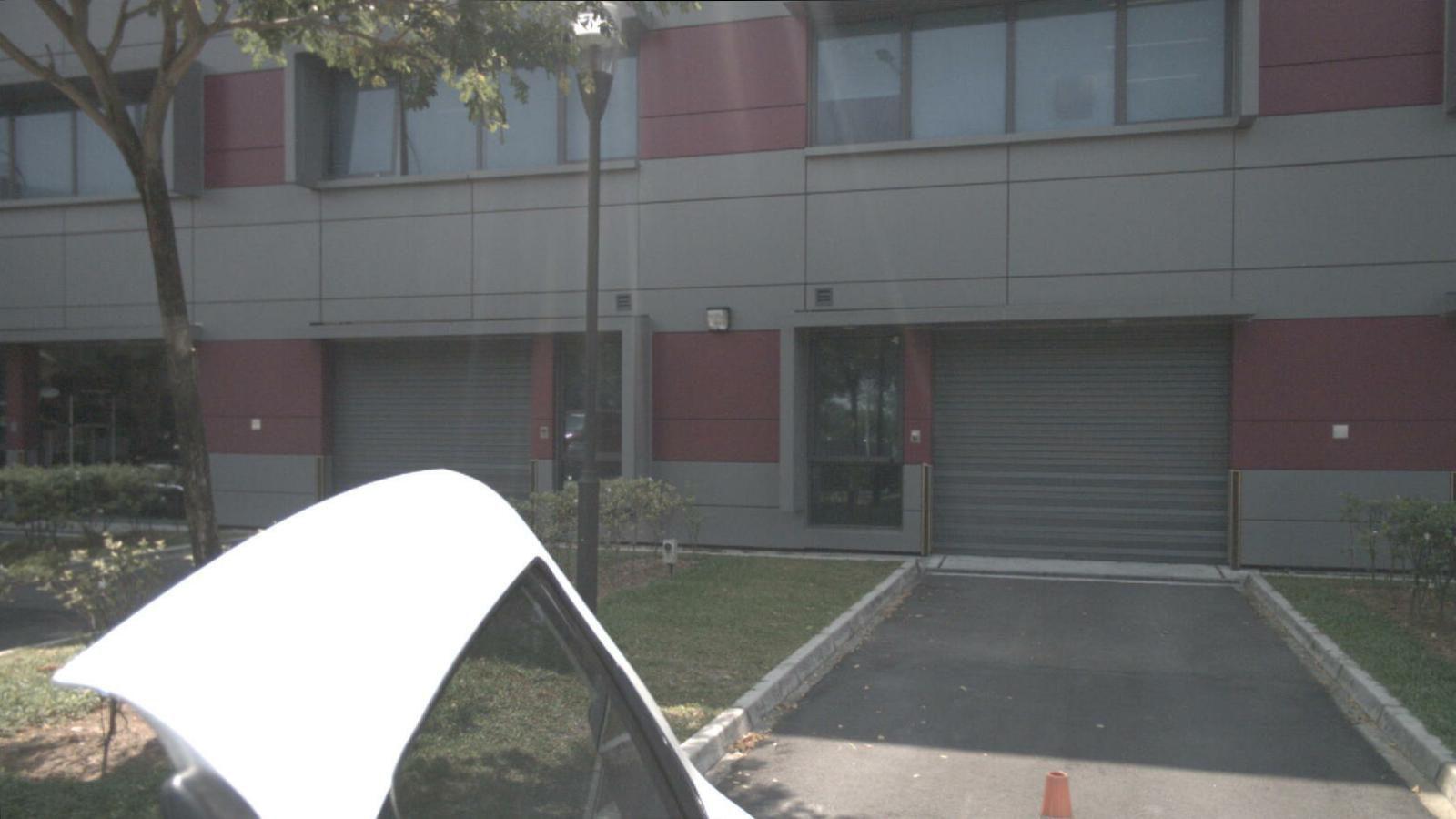}
        \end{subfigure}
        \hfill
        \begin{subfigure}[b]{0.32\textwidth}
            \includegraphics[width=\textwidth,height=2.6cm]{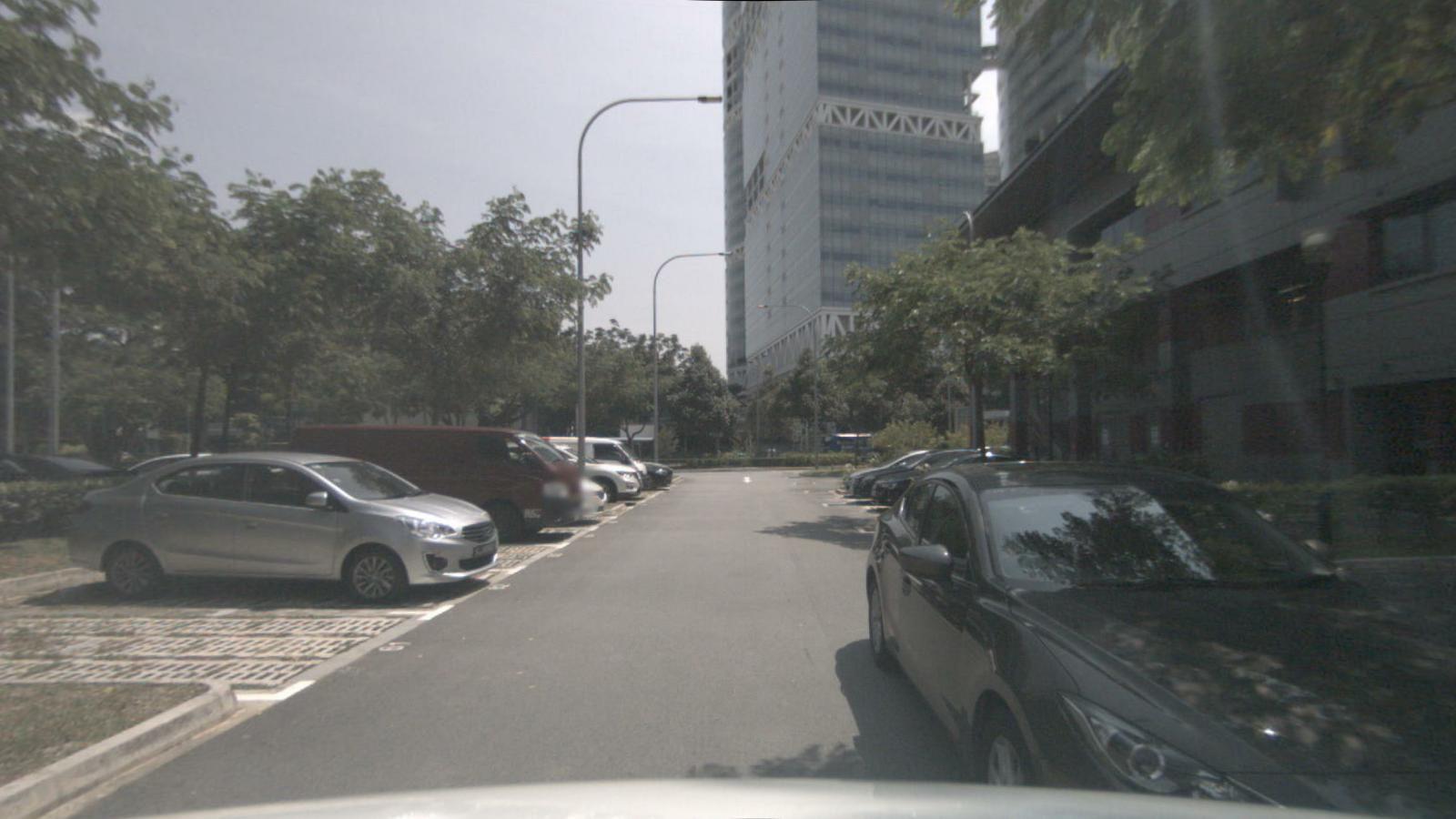}
        \end{subfigure}
        \hfill
        \begin{subfigure}[b]{0.32\textwidth}
            \includegraphics[width=\textwidth,height=2.6cm]{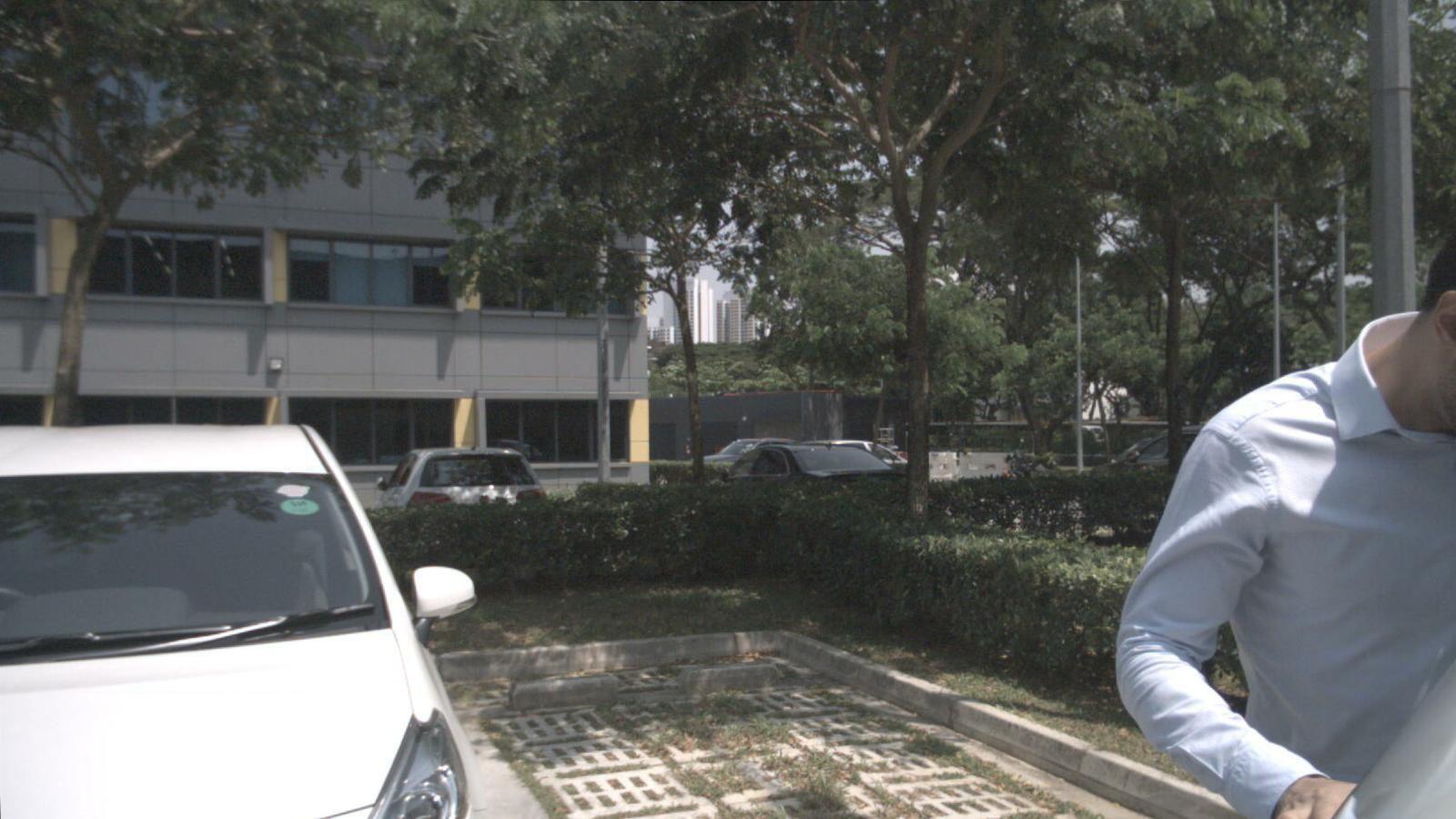}
        \end{subfigure}
    \end{minipage}
    \caption*{Camera Views.}
\end{subfigure}
    \end{minipage}

    \centering
    \begin{tcolorbox}[
        colback=white,
        colframe=black,
        width=\textwidth,
        boxrule=0.5mm,
        title={\textbf{Llama3.2-11B Reasoning Comparison: Parking Area Scenario}}
    ]
    \begin{minipage}{\textwidth}
        \small\textit{Ground Truth: Brake gently to come to a stop.}
    \end{minipage}
    
    \vspace{0.3cm}
    \small
    \begin{tabular}{p{3cm}|p{8cm}|p{4cm}}
        \small
        \textbf{Variant} & \textbf{Reasoning Process} & \textbf{Final Action} \\
        \hline
        Fine-tune w/ reasoning (reason) & ``The environment appears to be a parking area with parked vehicles and pedestrians nearby, indicating a potential for sudden movement. Keeping the vehicle stationary is crucial to avoid accidents [...] gently braking to a stop minimizes the risk of causing a rear-end collision [...] allows the vehicle to be prepared to react to any sudden changes.'' & ``Brake gently to a stop.'' \\
        \hline
        Fine-tuned w/o reasoning (simple) & ``The action taken by the ego vehicle is to keep going at the same speed. The reason is that there is no safety issue, and the ego vehicle is not moving.'' & ``Keep going at the same speed.'' \\
        \hline
        Instruct & ``To navigate the scene safely, the ego vehicle must carefully consider the surroundings [...] The presence of a white car and a black car [...] the presence of a sidewalk [...] suggests that pedestrians may be present [...] the safest actions would be: To slow down and be prepared to stop if necessary...'' & ``The safest actions for the ego vehicle would be to slow down, maintain a safe distance from other vehicles, and be aware of pedestrians on the sidewalk.'' \\
    \end{tabular}
    \end{tcolorbox}
    \caption{Comparing reasoning approaches within the Llama3.2-11B model family for a parking area scenario. The reason variant correctly identifies the need to stop completely through structured reasoning about the parking environment and pedestrian safety. The simple variant incorrectly assumes no safety issues and recommends maintaining speed with minimal reasoning. The Instruct variant provides general cautionary advice without the specific recommendation to stop completely.}
    \label{fig:llama_reasoning_comparison_parking}
\end{figure*}

\section{Experiments}

\subsection{Dataset}
We used a subset of the DriveLM dataset consisting of 5280 QA pairs including 800 frames with six camera views (front, front-left, front-right, back, back-left, and back-right) per frame from nuScenes dataset~\cite{caesar2020nuscenes}. For each frame, we generated structured reasoning chains using GPT-4o as described in our methodology. 

\subsection{Supervised Fine-Tuning}
We conducted experiments with three model families: Llama 3.2 vision (11B parameters)~\cite{LLama}, Llava 1.5 (7B parameters)~\cite{LLAVA}, and Qwen 2.5 VL (in both 3B and 7B variants)~\cite{Qwen25VL}. For each model family, we created three variants to compare different training approaches. The \textbf{reasoning-based fine-tuning (reason)} models were trained on examples that include structured reasoning chains before the answer, learning to generate explicit reasoning followed by the final answer. The \textbf{answer-only fine-tuning (simple)} models were trained on the same examples but without the reasoning component, tasked with directly producing answers without intermediate reasoning. The \textbf{baseline (Instruct)} models were the original instruction-tuned versions without domain-specific fine-tuning.

For all fine-tuning runs, we used the 8-bit quantized paged AdamW optimizer with a learning rate of 2e-4, weight decay of 0.01, and a linear learning rate schedule with 50 warmup steps. Training was conducted for 3 epochs on an NVIDIA H200 GPU using 4-bit quantization with mixed precision. We employed LoRA~\cite{LoRA} fine-tuning with rank 8, alpha 8, and no dropout, selectively fine-tuning language layers and MLP modules while freezing vision layers and attention modules. To manage memory constraints, we used a batch size of 1 with gradient accumulation steps of 12, gradient checkpointing, and early stopping with a patience of 2 evaluation steps.

\subsection{Evaluation}
We evaluated our models using the DriveLM evaluation benchmark, which provides a comprehensive assessment of model performance on autonomous driving tasks. Our evaluation metrics include:

\begin{itemize}
    \item \textbf{Accuracy}: Direct match with ground truth answers.
    \item \textbf{ChatGPT}: Evaluation of response quality using GPT-3.5 Turbo as a judge.
    \item \textbf{Match}: Semantic similarity between model predictions and ground truth.
    \item \textbf{BLEU-1,2,3,4}: N-gram precision metrics for text generation quality.
    \item \textbf{ROUGE-L}: Longest common subsequence-based metric for text similarity.
    \item \textbf{CIDEr}: Consensus-based metric for image description quality.
    \item \textbf{Final Score}: A weighted combination of accuracy, ChatGPT and Language evaluation, and match metrics.
\end{itemize}

These metrics allow us to assess both the accuracy of the driving decisions and the quality of the generated reasoning, providing a holistic view of model performance across different training approaches.

\begin{table*}[t]
\centering
\caption{Evaluation results for models with different fine-tuning approaches on DriveLM}
\label{tab:model_evaluation}
\renewcommand{\arraystretch}{1.2}
\resizebox{\textwidth}{!}{%
\begin{tabular}{l|cc|c|cccc|cc||c}
\toprule
\textbf{Model} & \textbf{Accuracy} & \textbf{ChatGPT} & \textbf{Match} & \textbf{Bleu\_1} & \textbf{Bleu\_2} & \textbf{Bleu\_3} & \textbf{Bleu\_4} & \textbf{ROUGE\_L} & \textbf{CIDEr} & \textbf{Final Score} \\
\midrule
\multicolumn{11}{l}{\textbf{Not fine-tuned (Instruct)}} \\
Qwen2.5-3B-Instruct & 0.00 & 0.73 & 0.26 & 0.02 & 0.01 & 0.01 & 0.01 & 0.12 & 0.00 & 0.35 \\
Qwen2.5-7B-Instruct & 0.00 & 0.62 & 0.25 & 0.00 & 0.00 & 0.00 & 0.00 & 0.10 & 0.00 & 0.30 \\
Llava1.5-7B-Instuct & 0.00 & 0.54 & 0.11 & 0.02 & 0.01 & 0.01 & 0.01 & 0.12 & 0.00 & 0.25 \\
Llama3.2-11B-Instruct & 0.00 & 0.71 & 0.22 & 0.01 & 0.01 & 0.01 & 0.01 & 0.16 & 0.00 & 0.34 \\
\midrule
\multicolumn{11}{l}{\textbf{Fine-tuned without reasoning}} \\
Qwen2.5-3B-simple & 0.32 & 0.53 & 0.43 & 0.02 & 0.01 & 0.01 & 0.01 & 0.17 & 0.00 & 0.37 \\
Qwen2.5-7B-simple & 0.32 & 0.46 & 0.34 & 0.01 & 0.01 & 0.01 & 0.01 & 0.13 & 0.01 & 0.33 \\
Llava1.5-7B-simple & 0.61 & 0.63 & 0.37 & 0.75 & 0.68 & 0.62 & 0.55 & 0.71 & 0.18 & 0.54 \\
Llama3.2-11B-simple & 0.26 & 0.48 & 0.38 & 0.10 & 0.07 & 0.05 & 0.04 & 0.22 & 0.00 & 0.34 \\
\midrule
\multicolumn{11}{l}{\textbf{Fine-tuned with reasoning}} \\
Qwen2.5-3B-reason & 0.47 & 0.56 & 0.37 & 0.46 & 0.38 & 0.32 & 0.26 & 0.55 & 0.11 & 0.45 \\
Qwen2.5-7B-reason & 0.67 & 0.62 & 0.36 & 0.70 & 0.64 & 0.58 & 0.52 & 0.69 & 0.18 & 0.54 \\
Llava1.5-7B-reason & 0.66 & 0.56 & 0.33 & 0.70 & 0.64 & 0.58 & 0.52 & 0.68 & 0.16 & 0.51 \\
Llama3.2-11B-reason & 0.68 & 0.62 & 0.36 & 0.73 & 0.67 & 0.60 & 0.54 & 0.70 & 0.17 & 0.55 \\
\bottomrule
\end{tabular}%
}
\end{table*}

\begin{figure*}[ht]
    \centering

    \begin{minipage}{\textwidth}
        \centering
        \begin{subfigure}[b]{\textwidth}
    \centering
    \begin{minipage}{\textwidth}
        \centering
        \begin{subfigure}[b]{0.32\textwidth}
            \includegraphics[width=\textwidth,height=2.6cm]{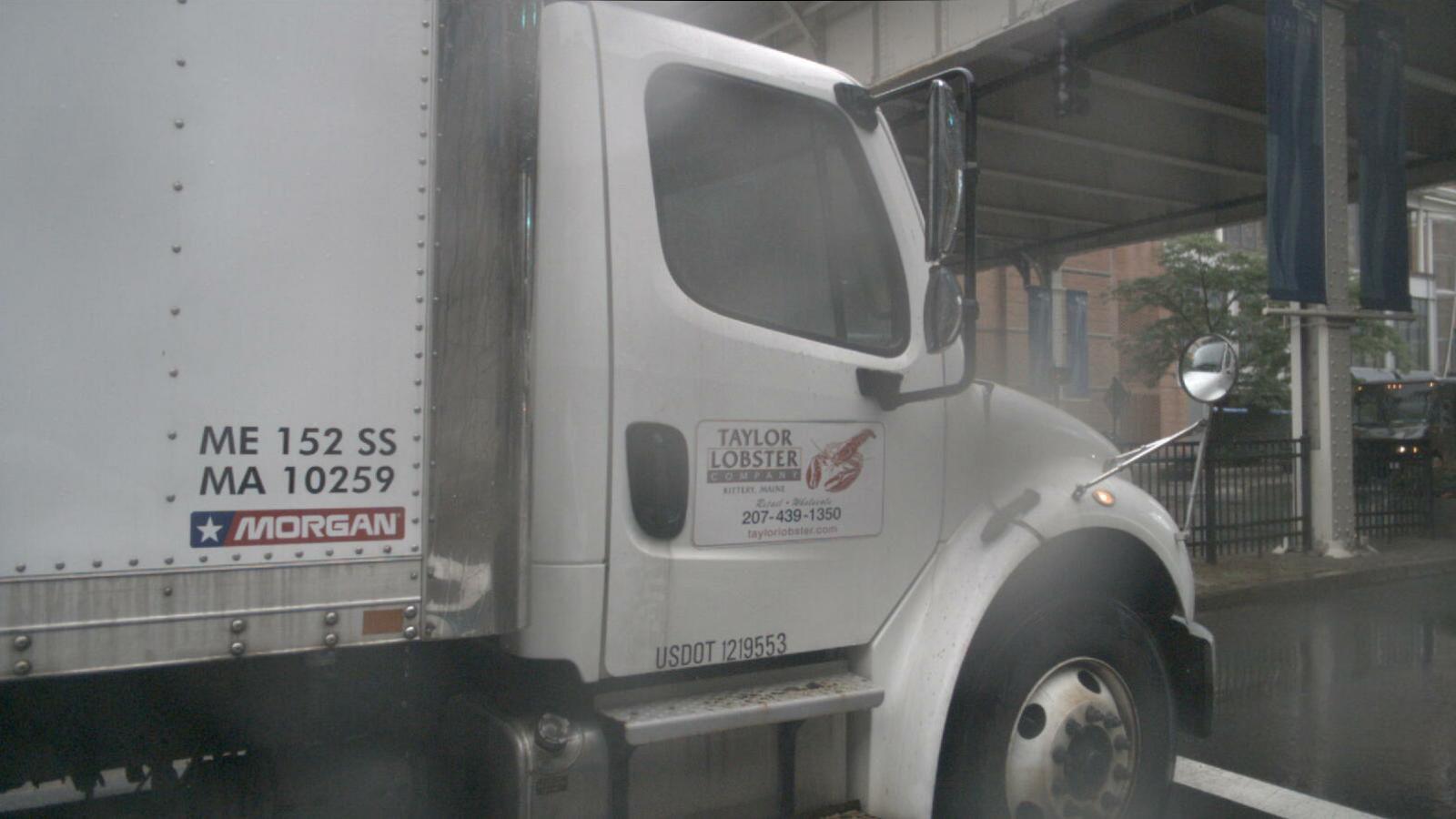}
        \end{subfigure}
        \hfill
        \begin{subfigure}[b]{0.32\textwidth}
            \includegraphics[width=\textwidth,height=2.6cm]{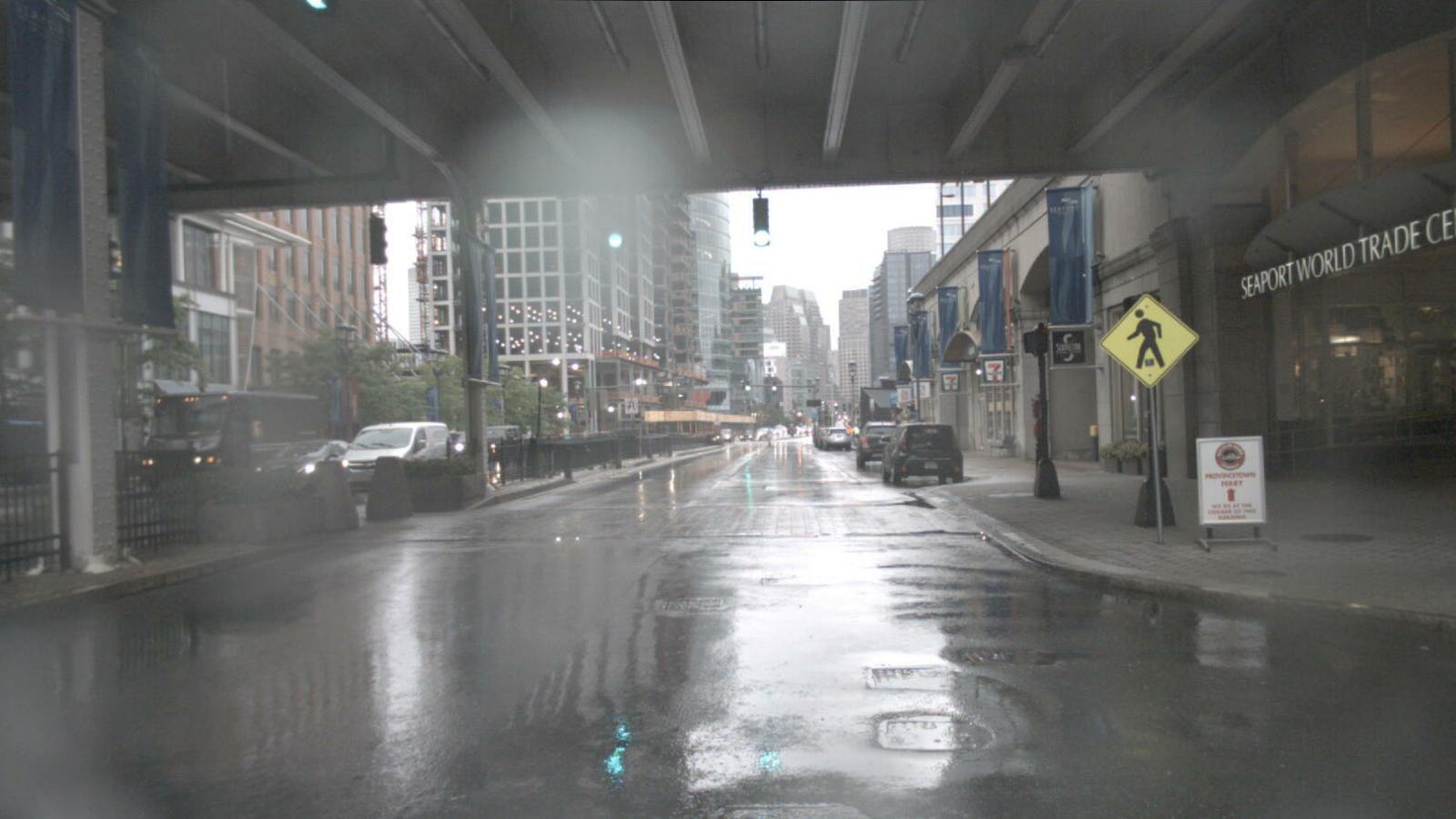}
        \end{subfigure}
        \hfill
        \begin{subfigure}[b]{0.32\textwidth}
            \includegraphics[width=\textwidth,height=2.6cm]{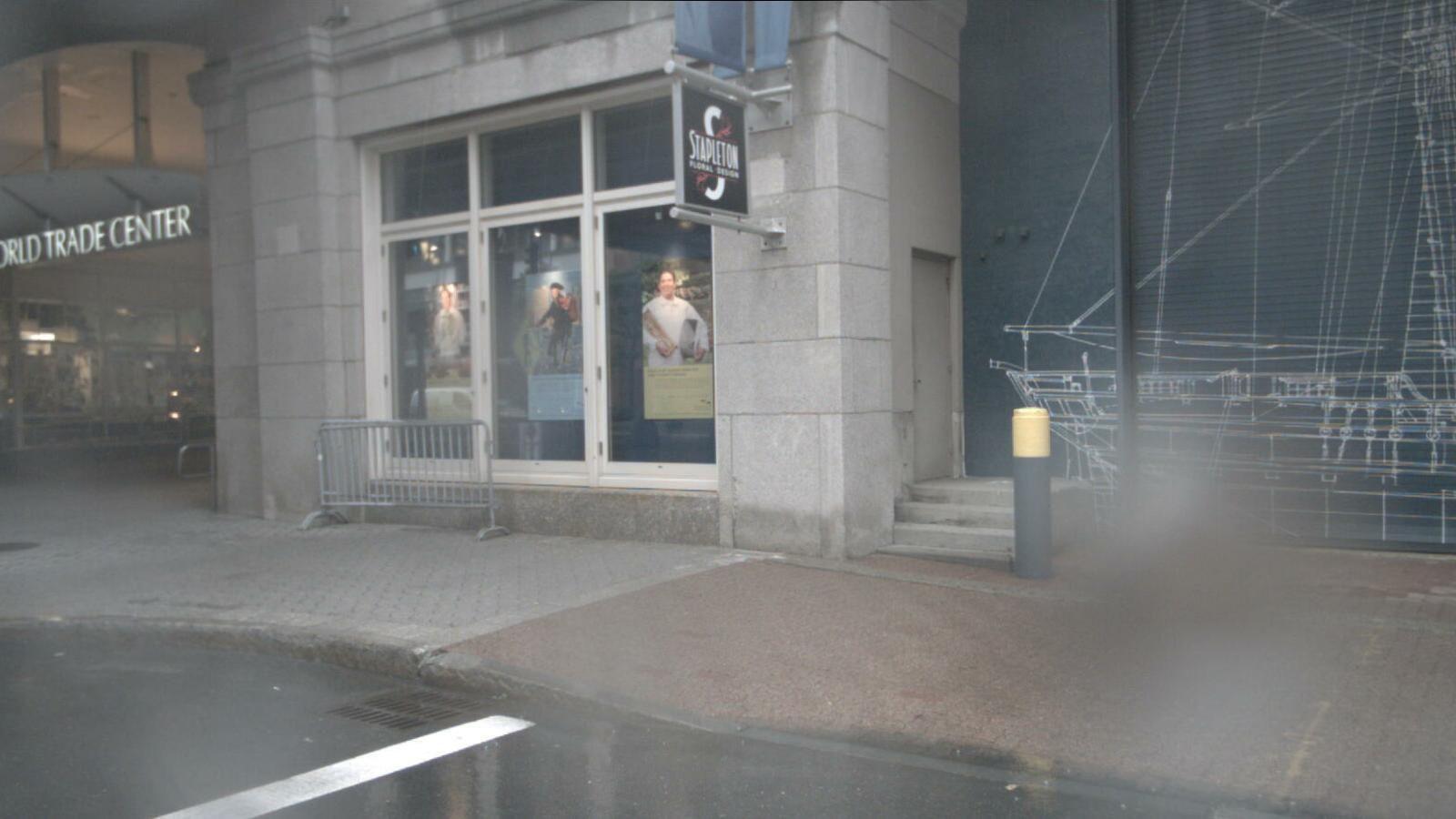}
        \end{subfigure}
        
        
        \begin{subfigure}[b]{0.32\textwidth}
            \includegraphics[width=\textwidth,height=2.6cm]{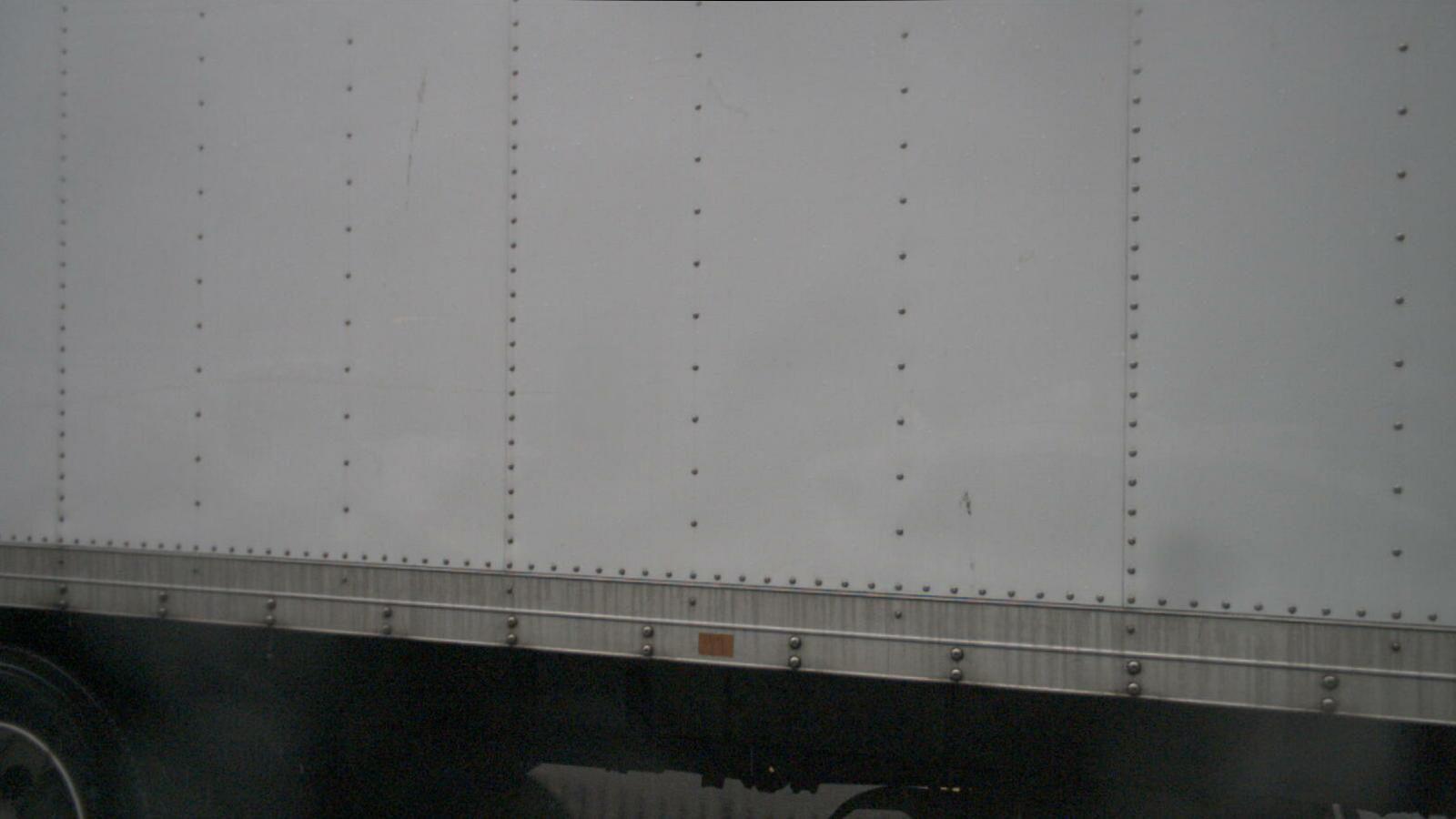}
        \end{subfigure}
        \hfill
        \begin{subfigure}[b]{0.32\textwidth}
            \includegraphics[width=\textwidth,height=2.6cm]{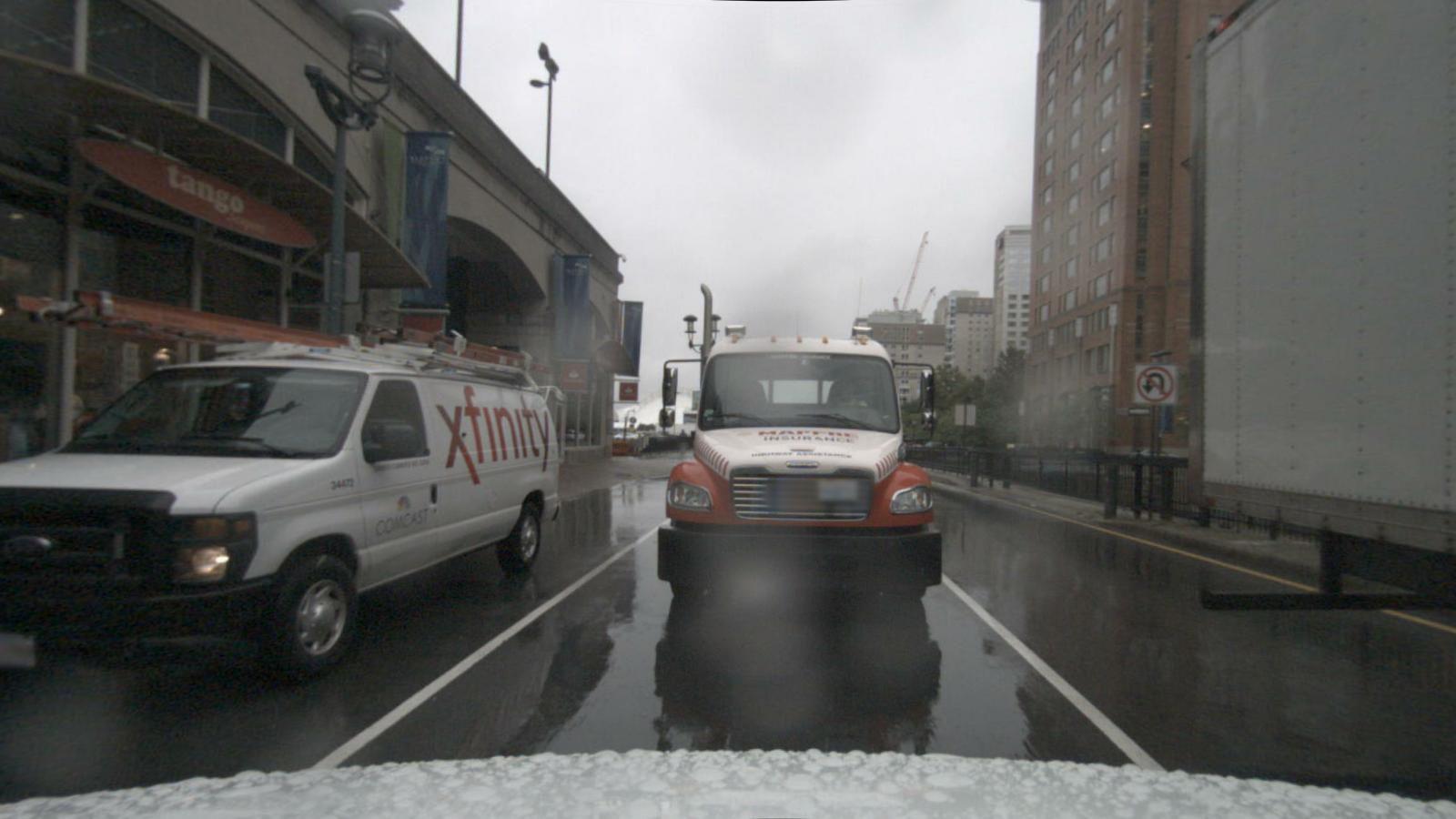}
        \end{subfigure}
        \hfill
        \begin{subfigure}[b]{0.32\textwidth}
            \includegraphics[width=\textwidth,height=2.6cm]{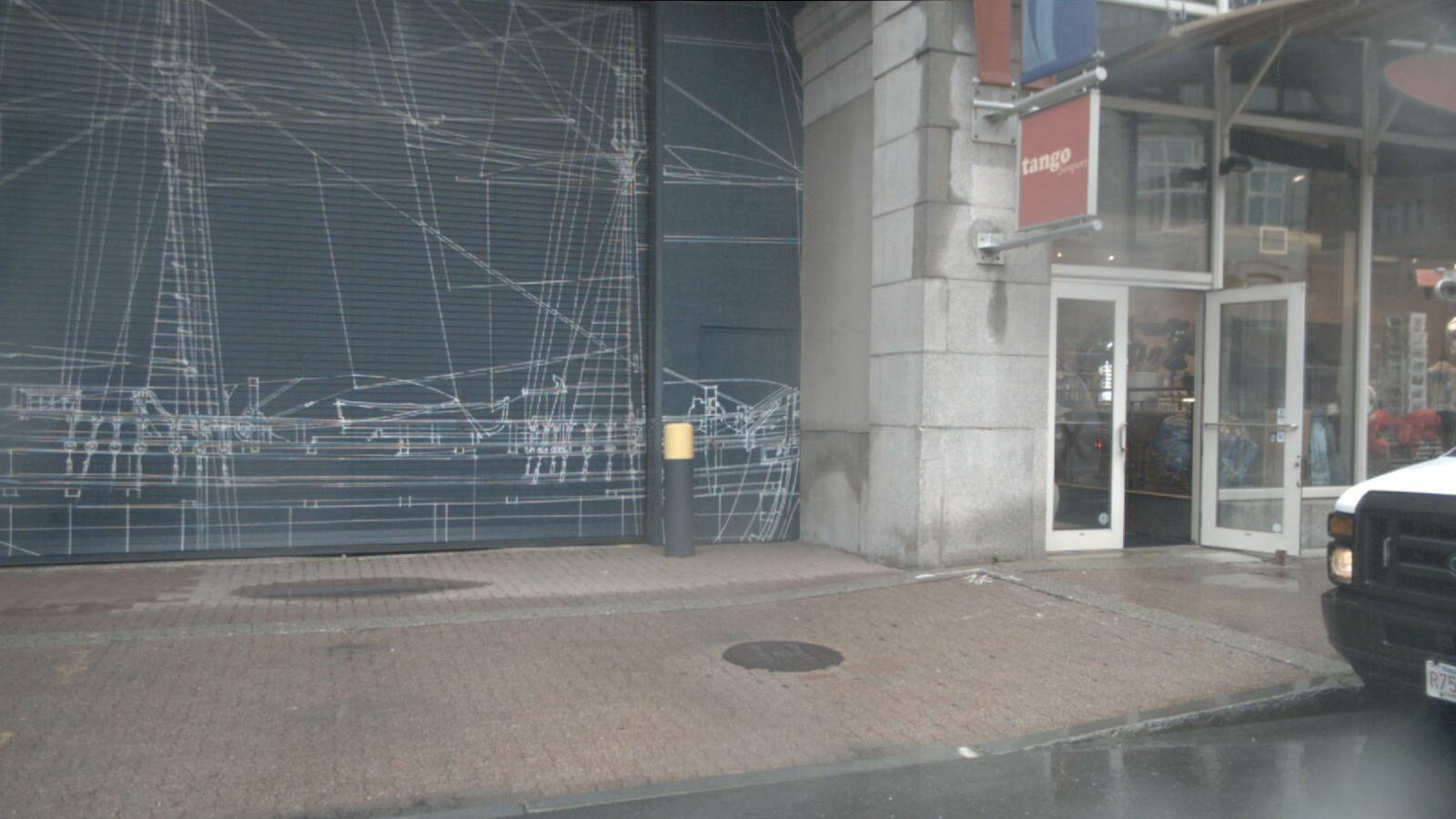}
        \end{subfigure}
    \end{minipage}
    \caption*{Camera Views.}
\end{subfigure}
    \end{minipage}

    \begin{tcolorbox}[
        colback=white,
        colframe=black,
        width=\textwidth,
        boxrule=0.5mm,
        title={\textbf{Scenario: Wet Road - Safe Actions for Ego Vehicle}}
    ]
    \begin{minipage}{\textwidth}
        \small\textit{Ground Truth: Decelerate gradually without braking; keep going at the same speed.}
    \end{minipage}
    
    \vspace{0.3cm}
    \begin{tabular}{p{2.3cm}|p{4cm}|p{4cm}|p{4.5cm}}
        \small
        \textbf{Model Family} & \textbf{reason} & \textbf{simple} & \textbf{Instruct} \\
        \hline
        Llama3.2-11B & ``Keep going at the same speed, decelerate gradually without braking.'' & ``Keep going at the same speed.'' & ``[...] The ego vehicle is situated at a four-way intersection with a stop sign [...] Yield to the Truck [...] Proceed with Caution [...]'' \\
        \hline
        Llava1.5-7B & ``Keep going at the same speed, decelerate gradually without braking.'' & ``Keep going at the same speed, decelerate gradually without braking, slightly offset to the left, and slightly offset to the right.'' & ``[Discusses navigation and awareness without specific action recommendation]'' \\
        \hline
        Qwen2.5VL-7B & ``Keep going at the same speed, decelerate gradually without braking.'' & ``[Confused response with multiple object IDs]'' & ``The ego vehicle should slow down, maintain a safe distance from other vehicles, be cautious of pedestrians, use headlights, and avoid sudden movements.'' \\
        \hline
        Qwen2.5VL-3B & ``Decelerate gradually, keep going at the same speed, and go ahead.'' & ``Brake gently to a stop.'' & ``The ego vehicle should maintain a safe speed, be prepared to stop if necessary [...]'' \\
    \end{tabular}
    \end{tcolorbox}
    \caption{Comparing model responses for safe driving actions in a wet road scenario. The reason models consistently recognize the need for gradual deceleration without braking on wet surfaces, while simple models vary in quality of response and Instruct models often add unnecessary actions or misinterpret the scene.}
    \label{fig:model_comparison_wet_construction}
\end{figure*}

\begin{figure*}[ht]
    \centering

    \begin{minipage}{\textwidth}
        \centering
        \begin{subfigure}[b]{\textwidth}
    \centering
    \begin{minipage}{\textwidth}
        \centering
        \begin{subfigure}[b]{0.32\textwidth}
            \includegraphics[width=\textwidth,height=2.6cm]{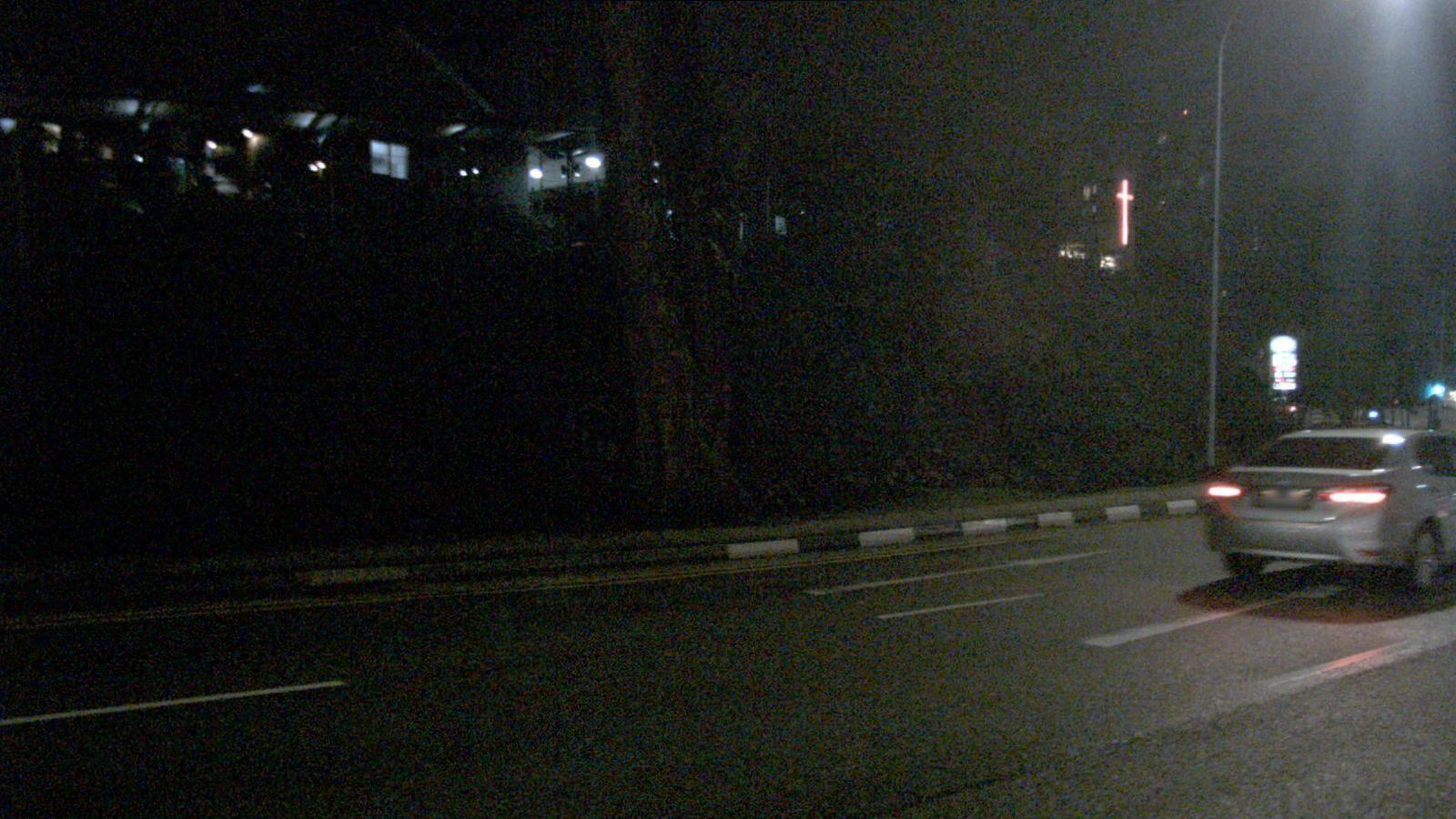}
        \end{subfigure}
        \hfill
        \begin{subfigure}[b]{0.32\textwidth}
            \includegraphics[width=\textwidth,height=2.6cm]{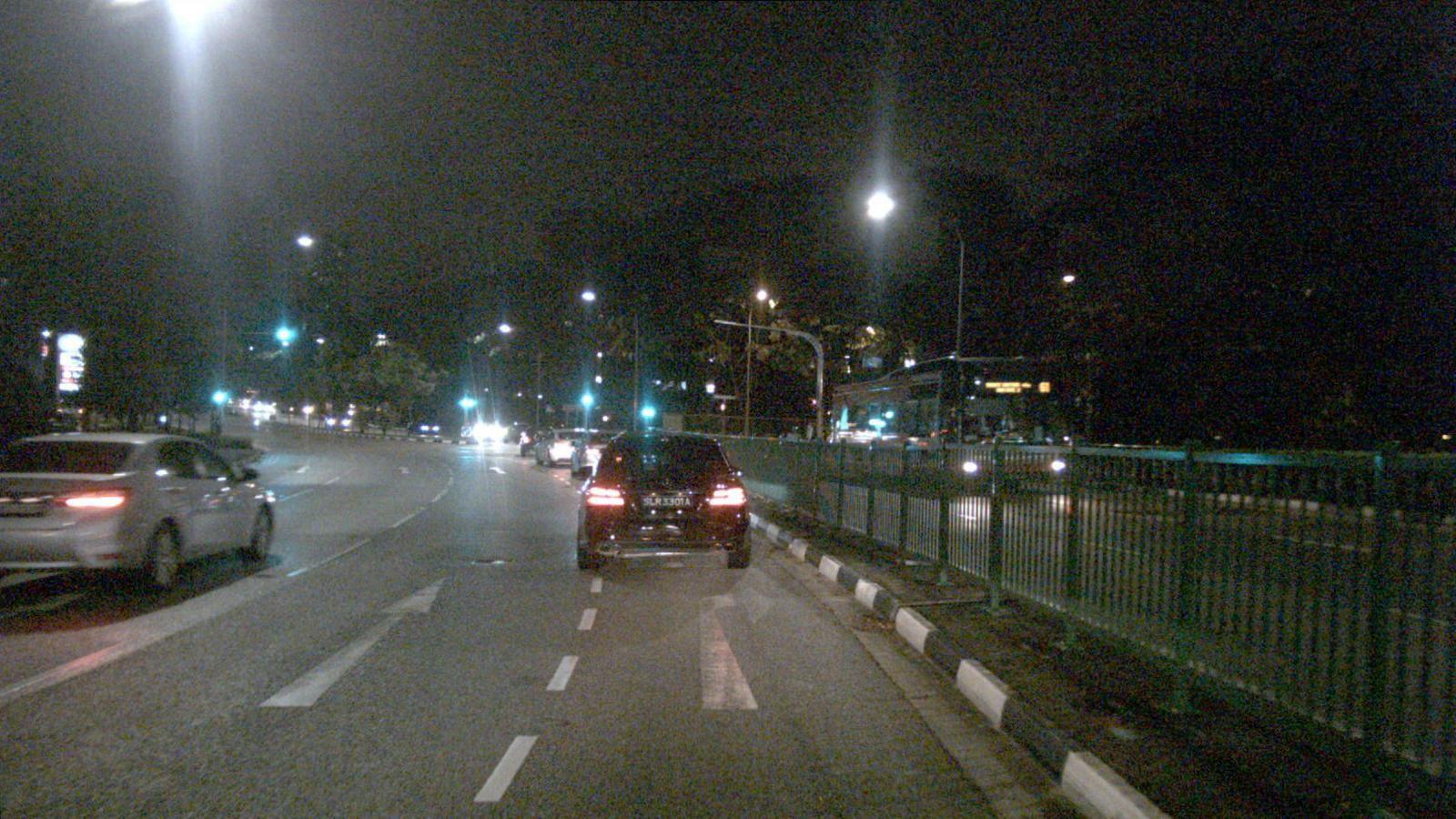}
        \end{subfigure}
        \hfill
        \begin{subfigure}[b]{0.32\textwidth}
            \includegraphics[width=\textwidth,height=2.6cm]{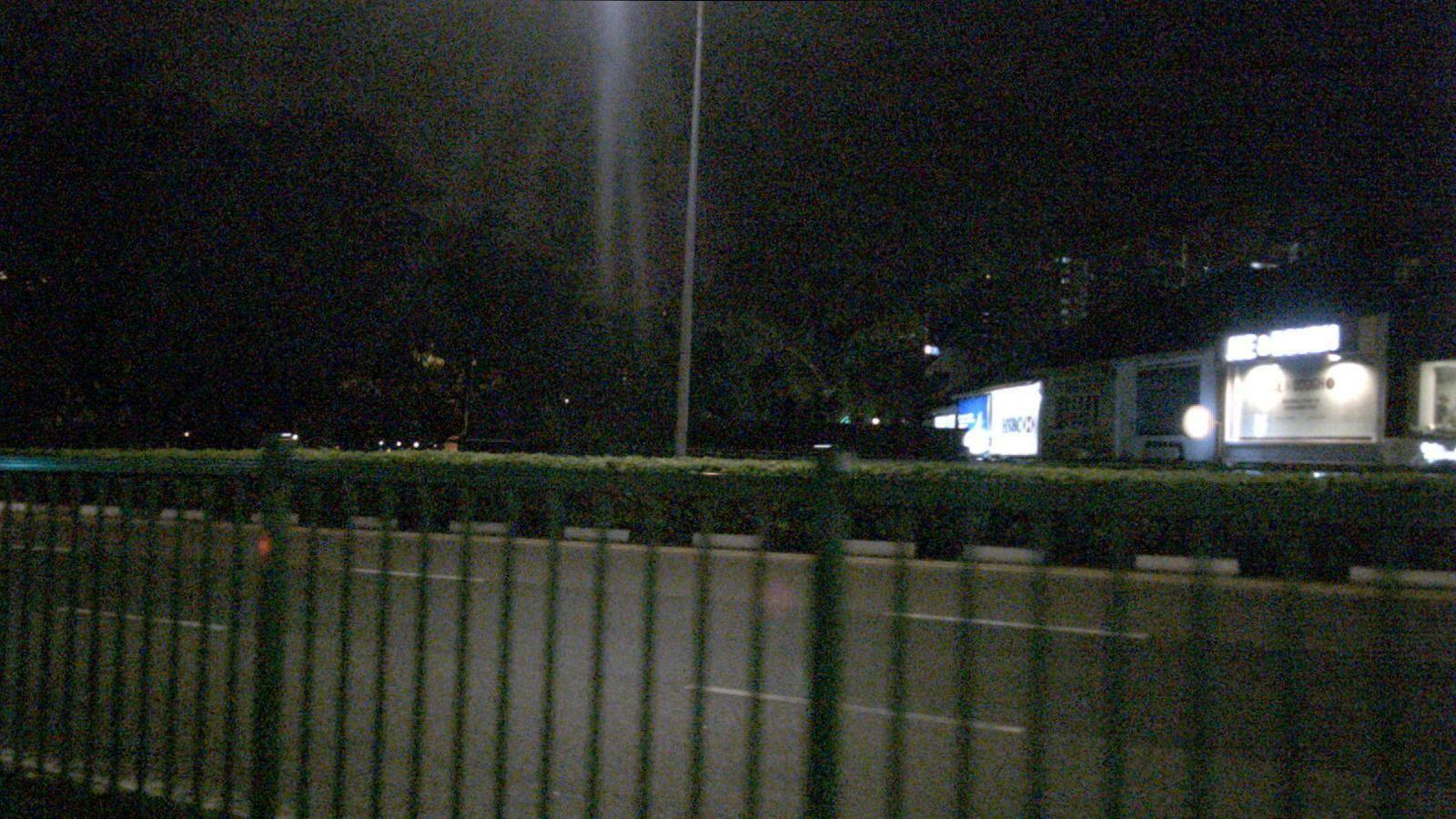}
        \end{subfigure}
        
        
        \begin{subfigure}[b]{0.32\textwidth}
            \includegraphics[width=\textwidth,height=2.6cm]{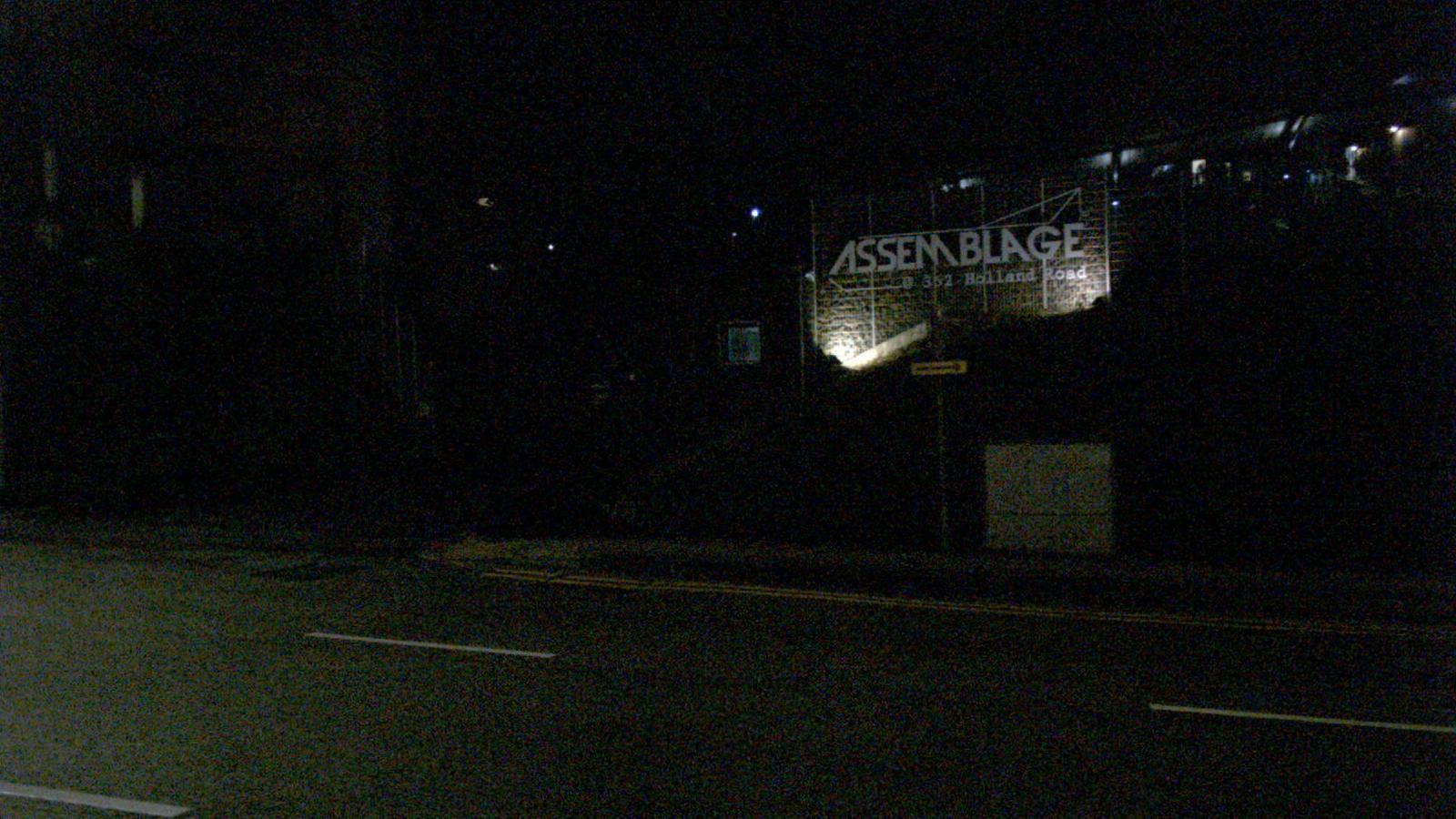}
        \end{subfigure}
        \hfill
        \begin{subfigure}[b]{0.32\textwidth}
            \includegraphics[width=\textwidth,height=2.6cm]{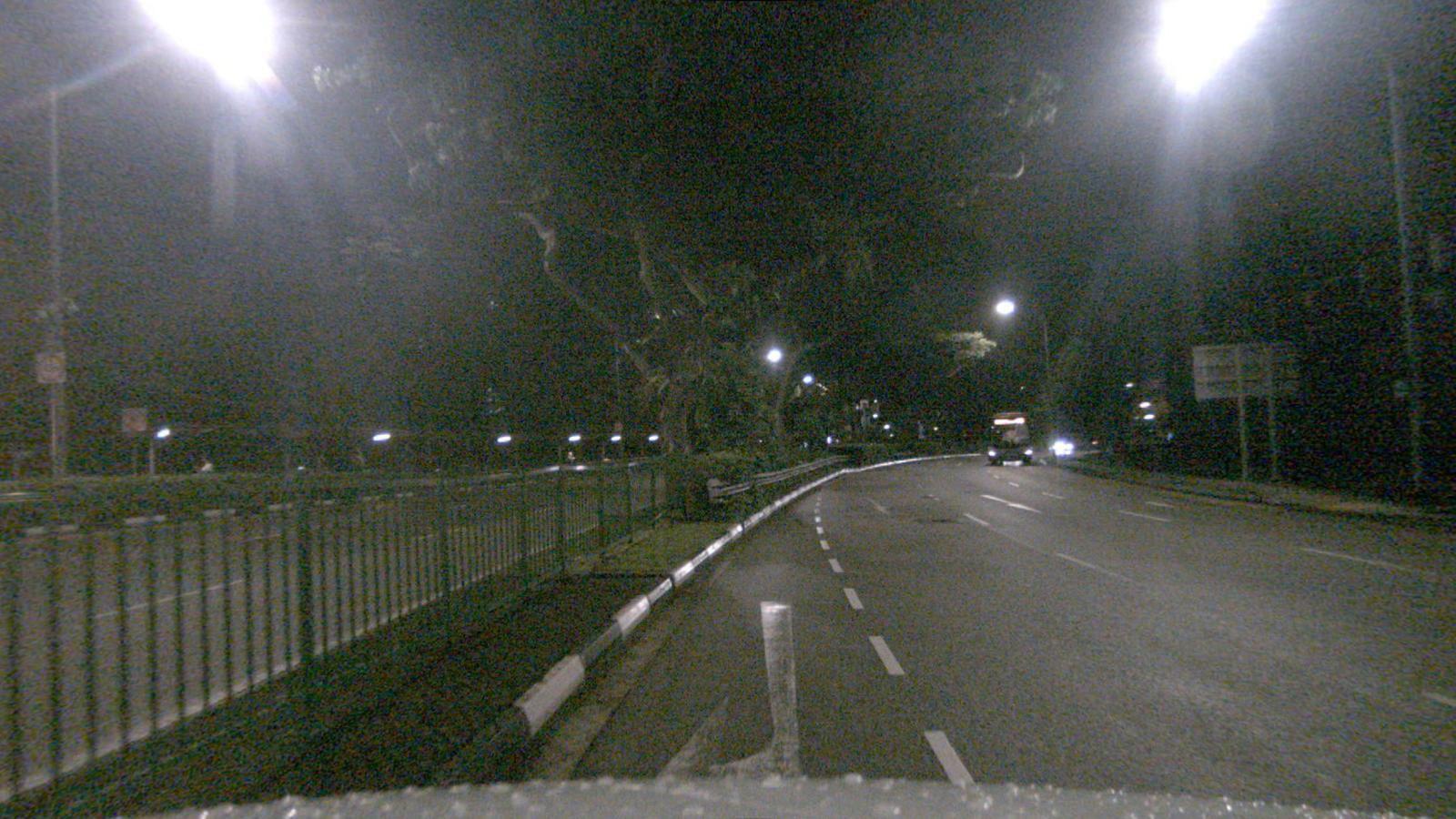}
        \end{subfigure}
        \hfill
        \begin{subfigure}[b]{0.32\textwidth}
            \includegraphics[width=\textwidth,height=2.6cm]{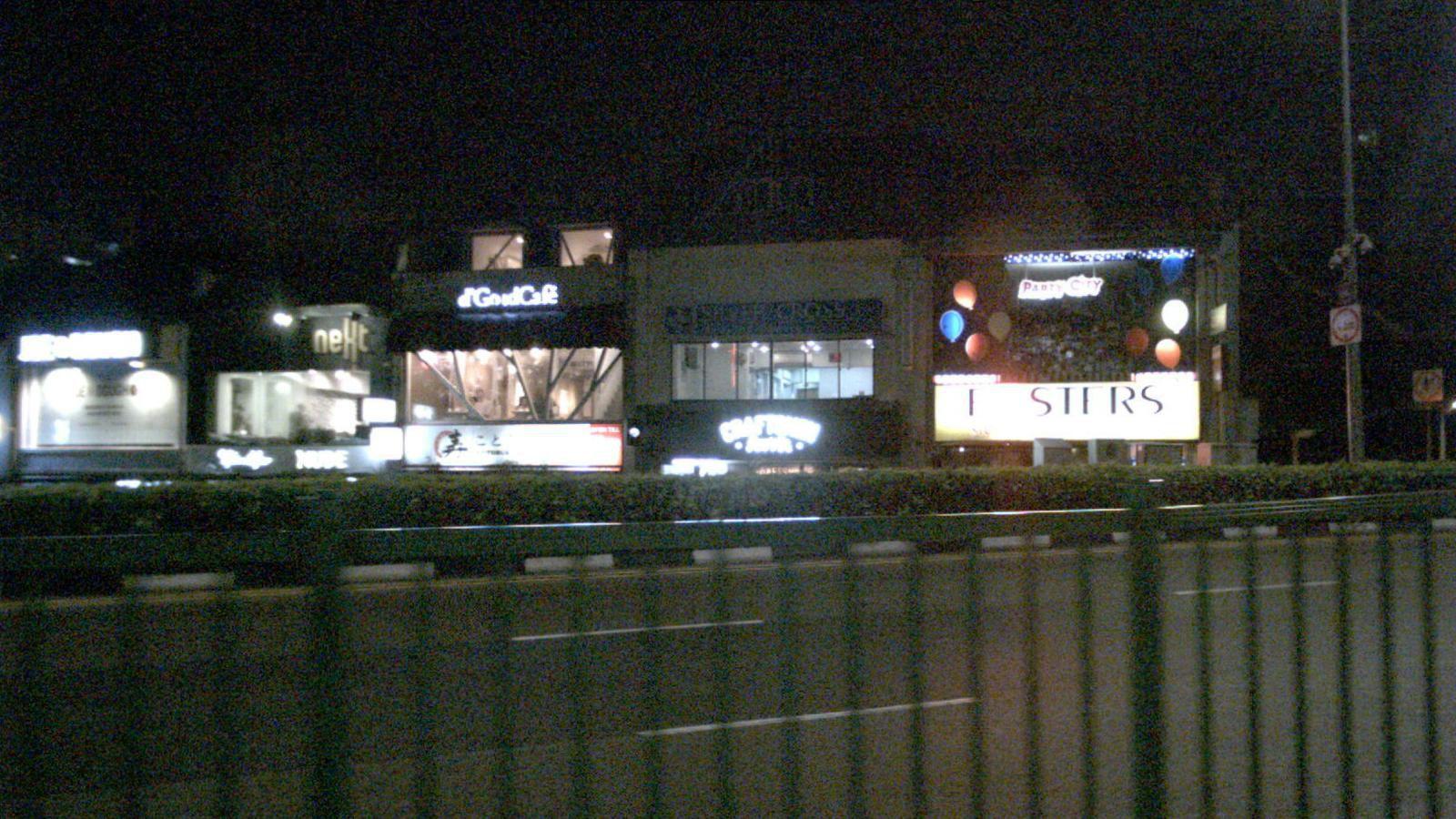}
        \end{subfigure}
    \end{minipage}
    \caption*{Camera Views.}
\end{subfigure}
    \end{minipage}

    \begin{tcolorbox}[
        colback=white,
        colframe=black,
        width=\textwidth,
        boxrule=0.5mm,
        title={\textbf{Scenario: Urban Nighttime Road - Safe Actions for Ego Vehicle}}
    ]
    \begin{minipage}{\textwidth}
        \small\textit{Ground Truth: Keep going at the same speed, decelerate gradually without braking.}
    \end{minipage}
    
    \vspace{0.3cm}
    \begin{tabular}{p{2.3cm}|p{4cm}|p{4cm}|p{4.5cm}}
        \small
        \textbf{Model Family} & \textbf{reason} & \textbf{simple} & \textbf{Instruct} \\
        \hline
        Llama3.2-11B & ``Keep going at the same speed, decelerate gradually without braking.'' & ``Keep going at the same speed.'' & ``[...] The ego vehicle should slow down, maintain a safe distance from other vehicles, and be aware of [...] pedestrians [...] The vehicle should also be prepared to stop if necessary [...].'' \\
        \hline
        Llava1.5-7B & ``Keep going at the same speed, decelerate gradually without braking.'' & ``Keep going at the same speed, decelerate gradually without braking.'' & ``[Discusses the driver being attentive without specific action recommendation]'' \\
        \hline
        Qwen2.5VL-7B & ``Keep going at the same speed, decelerate gradually without braking.'' & ``[Irrelevant object IDs and incomplete analysis]'' & ``The ego vehicle should slow down, maintain a safe distance from the car in front and be prepared to stop if necessary.'' \\
        \hline
        Qwen2.5VL-3B & ``Keep going at the same speed, slightly decelerate, and keep going straight.'' & ``Brake suddenly.'' & ``The ego vehicle should maintain a safe distance from the lead vehicle, be prepared to stop if necessary [...]'' \\
    \end{tabular}
    \end{tcolorbox}
    \caption{Comparing model responses for safe driving actions in a nighttime urban road scenario. The reason models consistently identify proper speed maintenance and gradual deceleration, while simple and Instruct variants show inconsistency and often recommend unnecessary braking that could be unsafe on wet roads.}
    \label{fig:model_comparison_night_scenario}
\end{figure*}

\section{Results}

Table~\ref{tab:model_evaluation} presents the comprehensive evaluation results of our experiments comparing different fine-tuning approaches across various model families. We observe several clear patterns that highlight the benefits of reasoning-based fine-tuning for driving scenario understanding.

\subsection{Impact of Reasoning-Based Fine-Tuning}

Models fine-tuned with reasoning (reason) consistently outperform their counterparts across nearly all metrics. The Llama3.2-11B-reason model achieves the highest overall performance with a final score of 0.55, followed closely by Qwen2.5-7B-reason (0.54) and Llava1.5-7B-reason (0.51). This pattern demonstrates that incorporating explicit reasoning during fine-tuning leads to more accurate and higher-quality outputs for driving decision tasks.

The most dramatic performance difference appears in the accuracy metric, where reasoning-based models achieve substantially higher scores (ranging from 0.47 to 0.68) compared to instruction-tuned models, which uniformly score 0.000. This indicates that fine-tuning with domain-specific data is crucial for driving scenario understanding, and reasoning-based fine-tuning further enhances this capability. Figure~\ref{fig:llama_reasoning_comparison_parking} illustrates an example of model responses to a planning question.

\subsection{Text Generation Quality}

The text generation quality metrics (BLEU and ROUGE\_L) reveal interesting insights into model capabilities. Reasoning-based models demonstrate superior performance in generating coherent and relevant text, with BLEU-4 scores ranging from 0.26 to 0.542 and ROUGE\_L scores between 0.55 and 0.70. In contrast, most models fine-tuned without reasoning or not fine-tuned at all achieve much lower scores on these metrics, with a notable exception being Llava1.5-7B-simple, which achieves competitive text generation quality (BLEU-4: 0.55, ROUGE\_L: 0.71).

The CIDEr scores, which measure consensus in image descriptions, follow a similar trend, with reason models achieving significantly higher scores (0.11-0.18) compared to most simple and instruct models.

\subsection{Role of Model Size and Architecture}

Our results indicate that model size and architecture play important roles in performance. Within the same training approach, larger models (11B and 7B parameter variants) generally outperform their smaller counterparts. For instance, Qwen2.5-7B-reason (0.54) significantly outperforms Qwen2.5-3B-reason (0.45).
 The Llama3.2-11B-reason model's superior performance indicates that larger language models with appropriate visual inputs can effectively reason about visual driving scenarios.

\subsection{ChatGPT Evaluation}

In the ChatGPT evaluation, which uses GPT-3.5 Turbo as a judge, we observe a different trend. The non-fine-tuned instruction models score highest on this metric (0.54-0.73), followed by fine-tuned models without reasoning (0.46-0.63) and then reasoning-based models (0.56-0.62). This suggests that instruction-tuned models may produce responses that appear more natural or aligned with GPT-3.5 Turbo's judgment criteria, even if they are less accurate for the specific driving tasks.

\subsection{Performance Trade-offs}

Our results reveal important trade-offs between different approaches. While reasoning-based fine-tuning excels in accuracy and text quality metrics, it slightly underperforms in the ChatGPT evaluation compared to instruction-tuned models. This suggests that different fine-tuning approaches affects different aspects of model performance, and the choice of approach should depend on the specific requirements of the deployment scenario.

The exceptional performance of Llava1.5-7B-simple (0.54) compared to other simple fine-tuned models highlights that some architectures may be particularly well-suited for certain tasks even without explicit reasoning components.

\subsection{Combined Performance}

The final score, which combines accuracy, ChatGPT evaluation, and match metrics, provides a holistic view of model performance. Reasoning-based fine-tuning consistently leads to the highest overall performance, with all reason models outperforming most of their simple and instruct counterparts. The only exception is Llava1.5-7B-simple, which achieves a competitive final score comparable to reasoning-based models.

In summary, our results demonstrate that reasoning-based fine-tuning significantly enhances model performance on driving decision tasks, particularly in terms of accuracy and text generation quality. The benefits of this approach are consistent across different model architectures and sizes, though the magnitude of improvement varies. Figures~\ref{fig:model_comparison_wet_construction} and~\ref{fig:model_comparison_night_scenario} illustrate an example of model responses to a planning question in wet road condition and night time scenario.

\section{Conclusion}

We investigated the impact of explicit reasoning in vision-language models for autonomous driving decision-making. Our experiments with Llama3.2-11B, Llava1.5-7B, and Qwen2.5 models demonstrate that reasoning-based fine-tuning significantly enhances performance compared to answer-only fine-tuning or using instruction-tuned models without domain adaptation.

Our results reveal three key insights: (1) domain-specific fine-tuning is crucial for driving scenario understanding, as shown by the stark contrast in accuracy between fine-tuned and non-fine-tuned models; (2) reasoning during fine-tuning leads to more accurate answers and higher-quality text generation, with Llama3.2-11B-reason achieving the highest overall performance; and (3) while larger models generally perform better, properly fine-tuned smaller models can still achieve competitive results.

These findings have important implications for developing more reliable autonomous driving systems. By enabling VLMs to reason explicitly about driving scenarios, we address the black-box nature of deep learning models in safety-critical applications. Limitations of our work include the modest dataset size and reliance on text-based evaluation metrics. Future work should focus on scaling up the dataset, developing specialized safety-oriented metrics, and investigating how explicit reasoning affects model robustness to out-of-distribution scenarios.

\section{Acknowledgment}

Computational resources were provided in part through NSF MRI Award Number 2320600.



{
    \small
    \bibliographystyle{ieeenat_fullname}
    \bibliography{main}
}


\end{document}